%% file: main.tex
\documentclass[11pt]{article}
\usepackage{times}
\usepackage{subcaption}

\usepackage[utf8]{inputenc}
\usepackage[T1]{fontenc}
\usepackage{fullpage}
\usepackage{url}
\usepackage{textcomp}
\usepackage[colorlinks=true,allcolors=blue,urlcolor={black}]{hyperref}

\usepackage[numbers]{natbib}
\usepackage{authblk}
\usepackage{booktabs}
\usepackage{amsmath, amsfonts, amssymb, amsthm, amsbsy, amscd, bm, bbm,mathrsfs}

\usepackage{multicol}
\usepackage{multirow}

\usepackage{amsmath, amsfonts, amssymb, amsthm, amsbsy, amscd, bm, bbm,mathrsfs}
\usepackage{dsfont}
\usepackage{multicol}
\usepackage{multirow}
\usepackage{subcaption}
\usepackage{subfiles} 
\usepackage{wrapfig}
\usepackage{enumitem}
\usepackage{graphicx}
\graphicspath{{figure/}}

\usepackage{caption}
\usepackage[ruled,vlined]{algorithm2e}
\SetKw{kwInput}{Input:}
\SetKw{kwOutput}{Output:}

\usepackage{nccmath}

\newcommand{\wps}{w^{\text{PS}}}

\SetCommentSty{mycommfont}

\newcommand{\easgd}{\texttt{S-EASGD}\xspace}
\newcommand{\bmuf}{\texttt{S-BMUF}\xspace}
\newcommand{\allreduce}{\texttt{S-MA}\xspace}
\newcommand{\freasgd}{\texttt{FR-EASGD}\xspace}
\newcommand{\freasgdfive}{\texttt{FR-EASGD-5}\xspace}
\newcommand{\freasgdlarge}{\texttt{FR-EASGD-30}\xspace}
\newcommand{\frbmuf}{\texttt{FR-BMUF}\xspace}
\newcommand{\frma}{\texttt{FR-MA}\xspace}

\newcommand{\fbdata}[1]{\texttt{Dataset-#1}}
\newcommand{\shadowsync}{\textbf{\texttt{ShadowSync}}\xspace}
\newcommand{\norm}[1]{\|#1\|}

\newcommand{\eps}{\texttt{EPS} }

\newtheorem{definition}{Definition}[section]

\newcommand{\yhat}{\widehat{p}}

\usepackage{subfiles} 
\usepackage{wrapfig}
\usepackage{graphicx}
\usepackage{multicol}
\graphicspath{{figure/}}

\begin{document}

\title{\bf ShadowSync: Performing Synchronization in the Background for Highly Scalable Distributed Training}

\author[1]{Qinqing Zheng\thanks{These authors contributed to this work while they were working at Facebook.}}
\author[2]{Bor-Yiing Su}
\author[2]{Jiyan Yang}
\author[2]{Alisson Azzolini}
\author[3]{Qiang Wu$^*$}
\author[4]{Ou Jin$^*$}
\author[2]{Shri Karandikar}
\author[2]{Hagay Lupesko}
\author[2]{Liang Xiong}
\author[2]{Eric Zhou}
\affil[1]{University of Pennsylvania}
\affil[2]{Facebook}
\affil[3]{Horizon Robotics}
\affil[4]{Cruise}
\renewcommand\Authsep{\hskip10pt}
\renewcommand\Authand{\hskip10pt}
\renewcommand\Authands{\hskip10pt}
\date{February 20, 2021}                     
\renewcommand\Affilfont{\itshape}

\maketitle

\begin{abstract}
\input{00-abstract}
\end{abstract}

\input{01-introduction_new}

\input{02-related}

\input{03-system}

\input{033-algo}
\input{04-experiments}
\input{05-conclusion}

\subsection*{Acknowledgement}
We would like to thank Mohamed Fawzy, Ozgur Ozkan, Mike Rabbat, Chonglin Sun, Chenguang Xi and Tommy Yu for
helpful discussions and consistent support.

\bibliographystyle{abbrvnat}

\bibliography{main}
\end{document}

%% file: 00-abstract.tex
Recommendation systems are often trained with a tremendous amount of data, and distributed training is the workhorse to shorten the training time. While the training throughput can be increased by simply adding more workers,  it is also increasingly challenging to preserve the model quality. In this paper, we present \shadowsync, a distributed framework specifically tailored to modern scale recommendation system training. In contrast to previous works where synchronization happens as part of the training process, \shadowsync separates the synchronization from training and runs it in the background. Such isolation significantly reduces the synchronization overhead and increases the synchronization frequency,  so that we are able to obtain both high throughput and excellent model quality when training at scale. The superiority of our procedure is confirmed by experiments on training deep neural networks for click-through-rate prediction tasks. Our framework is capable to express data parallelism and/or model parallelism, generic to host various types of synchronization algorithms, and readily applicable to large scale problems in other areas.

%% file: 01-introduction_new.tex
\section{Introduction}\label{sec:intro}
As one of the most important real-world applications of machine learning, recommendation
systems pervade our everyday lives. Personalized recommendations can be perceived in
many aspects, for example, Netflix selects movies for us; Spotify recommends
songs to us; Facebook personalizes news feed to us.  A key ingredient that drives the
success of personalized recommendation is data. To provide satisfactory personalized
experience to the users, recommendation systems are usually trained with tremendous
amount of data. Two types of information are typically utilized: the user-item
interaction histories, and user/item features (or their latent representation), where each type
of data might have massive volume. To scale up to modern industry scale of data,
distributed training becomes necessary.

The most common distributed training algorithms include the bulk-synchronous
implementations of stochastic gradient descent (SGD) \cite{robbins1951stochastic} and other variants
such as AdaGrad\cite{adagrad}, RMSProp\cite{rmsprop}, and ADAM\cite{adam}. In
essence, those algorithms expresses \emph{data parallelism} on a number of trainers at the batch
level: every trainer has a copy of the model; in each iteration, every trainer computes the gradient on its
local batch and then averages them to perform the update\cite{chu2007map, mann2009efficient}.
Nevertheless, as we drastically increase the amount of data as well as the complexity of our models,
those classic algorithms are not suitable for modern recommendation systems anymore. It is not uncommon to see models that simple
bulk-synchronous algorithms are hard to deal with, see, e.g., \cite{hogwild, distbelief}.

There are several major challenges for training modern recommendation systems. (i)
First, synchronous algorithms are often slow in practice due to the existence of
straggler machines, and asynchronous implementations become a must. However,
asynchronous algorithms often suffer from poor convergence \cite{chen2016revisiting}. (ii) Second, data
parallelism only is not sufficient due to the growth of model complexity, where a model
often exceeds the memory of a single host. As a result, we need to express \emph{model
parallelism} which partition the model into multiple hosts \cite{distbelief, parameterserver}. (iii) 
Third, to scale up the training, we often tend to add more workers and increase the batch size.
However, it is well known that large batch size can potentially hurt model quality \cite{fbimagenet}.
More importantly,
several type of tasks, including the important click-through-rate (CTR) prediction problems,
are prone to overfitting; and one-pass training is commonly used to prevent that \cite{zhou2018deep}.
With that, we cannot iterate the training data again and again to further optimize the model.
With this very strict setting, when we express data
parallelism on more workers, each worker processes less data. In order to keep sufficient number of updates for convergence, one might need to limit the batch size.  Therefore, increasing the throughput while preserving the model quality becomes extremely hard. 

This raises a pressing call for more sophisticated distributed training algorithms that
(1) deal with both model and data parallelism, (2) provide large throughput to finish
the training job in a reasonable amount of time, (3) and preserve model quality with
negligible accuracy loss. 

In this paper, we propose \shadowsync, a distributed training framework and associated synchronization methods
specifically tailored to modern industry scale recommendation systems that has the above appealing properties. 
Throughout this paper, we use the \emph{Deep Learning Recommendation Model} (DLRM)\cite{fb:dlrm}, one of the most widely adopted deep architecture for recommendation, as our target model to illustrate \shadowsync,
See Section~\ref{sec:optimization} for a brief review.

\shadowsync~allows us to express both model parallelism and data parallelism at the same time, and
different synchronization strategies are concurrently used for each of them. 
In addition to the carefully design mechanism, \shadowsync~is novel in its data parallelism synchronization algorithm. Unlike the classic algorithms that incorporate synchronization into the training (i.e., synchronization is essentially part of the training algorithm),
we completely separate synchronization from training and perform it in the background. 
In the language of software engineering, synchronization is executed by an additional thread in parallel with other training threads, without interfering or communicating with them. We call such thread the \emph{shadow} thread and hence term our approach \shadowsync.

Surprisingly, this simply disentanglement of synchronization and training is extremely powerful and effective in practice.
First, having synchronization as part of the training process introduced communication overhead. No matter synchronous or asynchronous methods we use, we need to stop the training while performing synchronization in the computing thread.
In fact, synchronization is usually the bottleneck of a training systems due to the stall of training; and there are much active research attention and efforts on reducing the synchronization overhead via techniques such as quantization, gradient sparsification, compression, and so on \cite{terngrad, deepcompression, onebit}. Since synchronization is isolated from training for the data parallelism part in \shadowsync, we are able to eliminate 
such overhead, thus keep both high sync frequency as well as high training throughput; see Section~\ref{sec:expr} for details.
Moreover, one major pain point of practical deployment in industry usage is the complexity and human efforts of maintaining, testing, and extending a software framework. Due to the well isolation of training and synchronization, we found \shadowsync~easy and flexible to use in practice. One can develop new synchronization algorithms that sync in the background, without touching any piece of the training implementation and vice versa, which makes it extremely convenient to test and maintain. Various algorithms are quickly developed and tested under this framework, including centralized and decentralized methods.

Although we primarily demonstrate our approach on DLRM, we emphasize \shadowsync~is a generic framework and readily extends to other models and regimes, wherever standard distributed optimization methods play a role. To summarize, our major contributions are as follows.
 \begin{itemize}[leftmargin=*]
     \item 
     We propose a distributed training framework \shadowsync~for modern recommendation system training. This framework is capable of
     expressing both model parallelism and data parallelism, or either one of them.
     For data parallelism, our framework synchronizes parameters in the background without interfering the training process, thus eliminate the synchronization overhead.
     The framework is generic to host various synchronization algorithms, 
     and we propose three under it: ShadowSync EASGD, a typical centralized asynchronous algorithm; Shadow BMUF and Shadow MA, two decentralized synchronous algorithms. 
     \item For practical implementation, the software abstraction of synchronization and training can be isolated.
     This makes our framework easy to maintain and test, flexible to future extensions, e.g., accommodating new algorithms.
     \item  We compare our algorithms with the foreground variants where synchronization is attached to training.  We empirically demonstrate that \shadowsync~enables us to keep
     high throughput and high sync frequency at the same time, 
     and hence achieves favorable model quality;
     whereas
     the foreground variants are sometimes bottlenecked by synchronization. 
     \item We compare the three ShadowSync algorithms mentioned above. We conclude that all of them have the same training throughput. ShadowSync EASGD has slightly better quality, and ShadowSync BMUF/MA consume fewer compute resources because their decentralized property: no need of the extra sync parameter servers.
 \end{itemize}

%% file: 02-related.tex
\section{Related Work}\label{sec:related}
As an interdisciplinary research area,
developing frameworks and algorithms for large scale distributed training
has received intensive attention from both optimization and system communities.

The research on distributed algorithms from the optimization perspective has been
mainly focusing on the data parallelism regime, with an emphasis on the convergence properties
and/or scalability of the proposed methods, see, e.g. \cite{lars, fbimagenet}.
Ealier works mainly discuss the parallelizing the gradient computations of SGD at the batch level (e.g. \cite{chu2007map, mann2009efficient}).
Those bulk-synchronous SGD has a few fundamental limitations: stragglers in the workers will slow down the synchronization, the barriers forced at the synchronization stage introduce extra overhead, 
and the failure in any worker will fail the whole training.
Those issues are overcame by the asynchronous SGD algorithms. Asynchronous SGD removes the dependency among the workers. It allows the workers use its local gradient to update the parameters, without aggregating the gradients
computed by the other workers.
The Hogwild! algorithm is a lock-free version of asynchronous SGD that can achieve a nearly optimal rate of convergence for certain problems \cite{hogwild}.
The Downpour SGD is another famous variant of asynchronous SGD
that adapted the parameter server and worker architecture \cite{distbelief}. 
Several other works studying asynchronous SGD include
\cite{chat2015async, zhao2016fast, dcasgd, de2017understanding, hakimi2019taming}. 

One pain point of parallelization at the gradient level is the expensive
communication overhead, especially when the shared parameter copy is hosted on a remote machine
(e.g. a parameter server). This motivated the researchers to propose new model synchronization
algorithms that let each worker owns a local replica of the model parameters, trains independently,
and periodically aggregate the local models.
The EASGD algorithm \cite{easgd} uses parameter servers to host a global copy of parameters,
which will aggregate with the local parameter replicas. 
Similar to the aforementioned gradient parallelization schemes, the synchronization of EASGD
can be done synchronously (every $k$ iterations) or asynchronous.
Instead of using parameter servers, the model averaging algorithm (MA) \cite{allreduce} 
utilizes the AllReduce primitive to aggregate the sub-models every $k$ iterations.
Contrasted with the EASGD algorithm, the network topology of MA is decentralized.
Other examples in this category include the blockwise model-update filtering (BMUF) algorithm \cite{bmuf},
the slow momentum algorithm\cite{wang2019slowmo} and so on.
In practice, the AllReduce primitive introduces huge synchronization overhead. The asynchronous decentralized parallel SGD \cite{adpsgd} and the stochastic gradient push \cite{sgp} algorithm rely on peer-to-peer communications among the workers, and will perform the gossiping-style synchronization.
For all of these algorithms, the synchronization always happens in the foreground. It is either 
incorporated in the backward pass, or added every $k$ iterations. 

From the system perspective, the research primarily focuses on the practical performance
of the distributed training system as a whole. Popular topics include the elastic scalability,
continuous fault tolerance, asynchronous data communication between nodes, and so on.
\cite{parameterserver} is one of the pioneer works that use parameter servers to host
the shared resources for large scale distributed machine learning problems.
One of the most famous framework utilizing parameter servers is DistBelief \cite{distbelief}, which introduces the concepts of model parallelism and data parallelism. DistBelief allows placing all the parameters on the parameter servers, and let the workers to use the Downpour SGD algorithm to read and update the shared parameters asynchronously.

%% file: 03-system.tex
\section{ShadowSync}\label{sec:sys}
We first give an overview of \shadowsync~in Section~\ref{sec:sys_overview}, using DLRM as our example model. 
Since we use both model and data parallelism, we discuss our optimization strategies specially tailored for each of them
in Section~\ref{sec:optimization}. 
Finally, we present the background synchronization algorithms in Section~\ref{sec:algo}.

\subsection{Framework Overview}\label{sec:sys_overview}
\begin{figure}[tb]
    \centering
    \includegraphics[width=0.6\columnwidth]{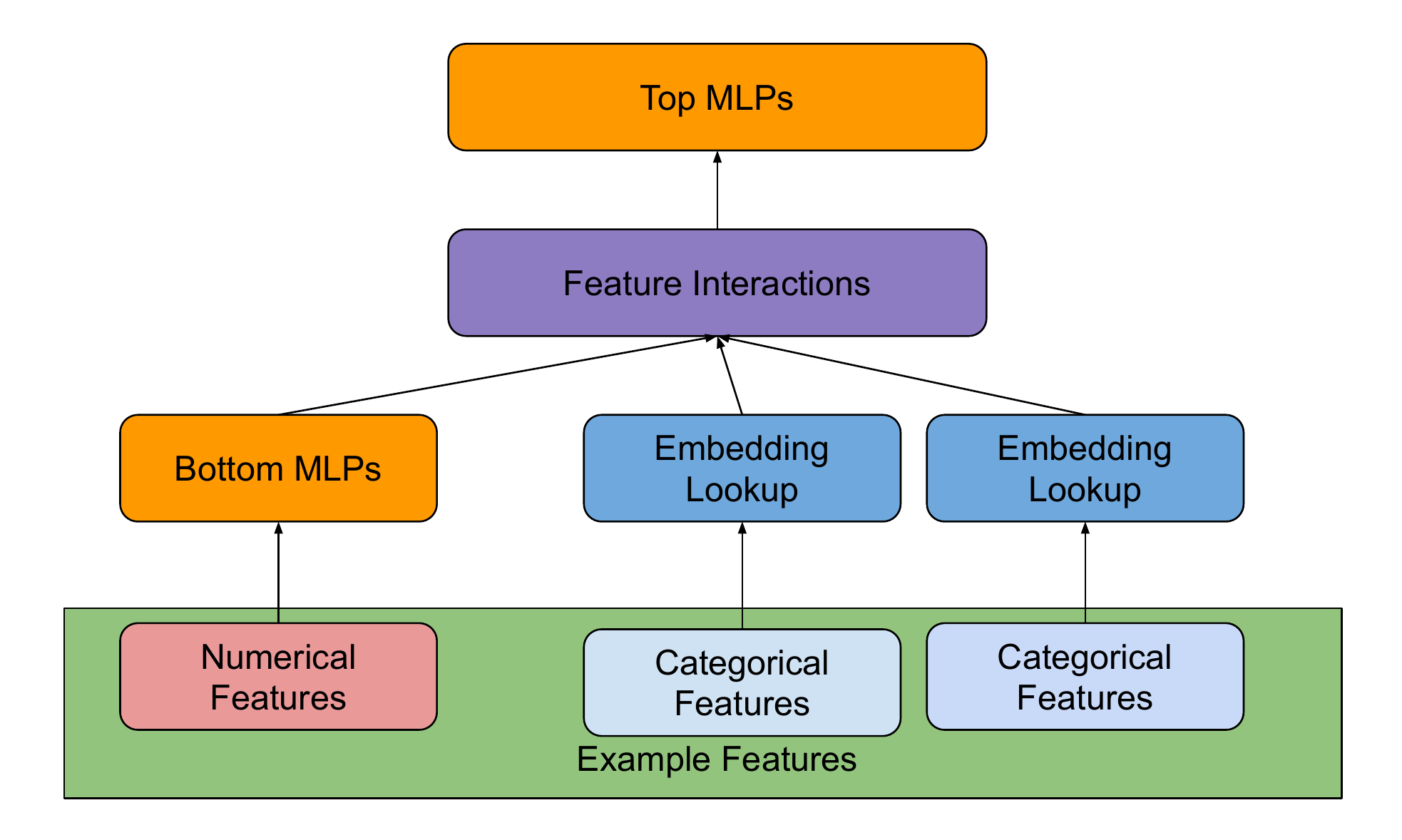}
    \vskip-5pt
    \caption{\small Overview of the DLRM \cite{fb:dlrm} model architecture.}
    \label{fig:dlrm}
\end{figure}
To demonstrate \shadowsync, we use DLRM \cite{fb:dlrm} as the primary example and focus training it throughout this paper. 
We hereby briefly review the DLRM architecture, as illustrated in Fig~\ref{fig:dlrm}.
The DLRM is composed of three components, from bottom to top: (i) feature transformation, (ii) feature interaction, and (iii) prediction network. The bottom feature transformation component contains the embedding look-up tables where the categorical features are transformed into latent embeddings, and multi-layer perceptrons (MLP) that transfers numerical features to latent presentations. This component captures the feature semantics and group features with similar semantics together.
The middle feature interaction component collect the transformed features, and generates helpful signals
through the interactions of them, e.g. the co-occurrence of them. The top prediction network is typically a MLP.
All of these strengths together makes the DLRM model expressive and capable of delivering high quality predictions. We refer readers to \citet{fb:dlrm} for the details.

\begin{figure}[tb]
    \centering
    \includegraphics[width=0.6\columnwidth]{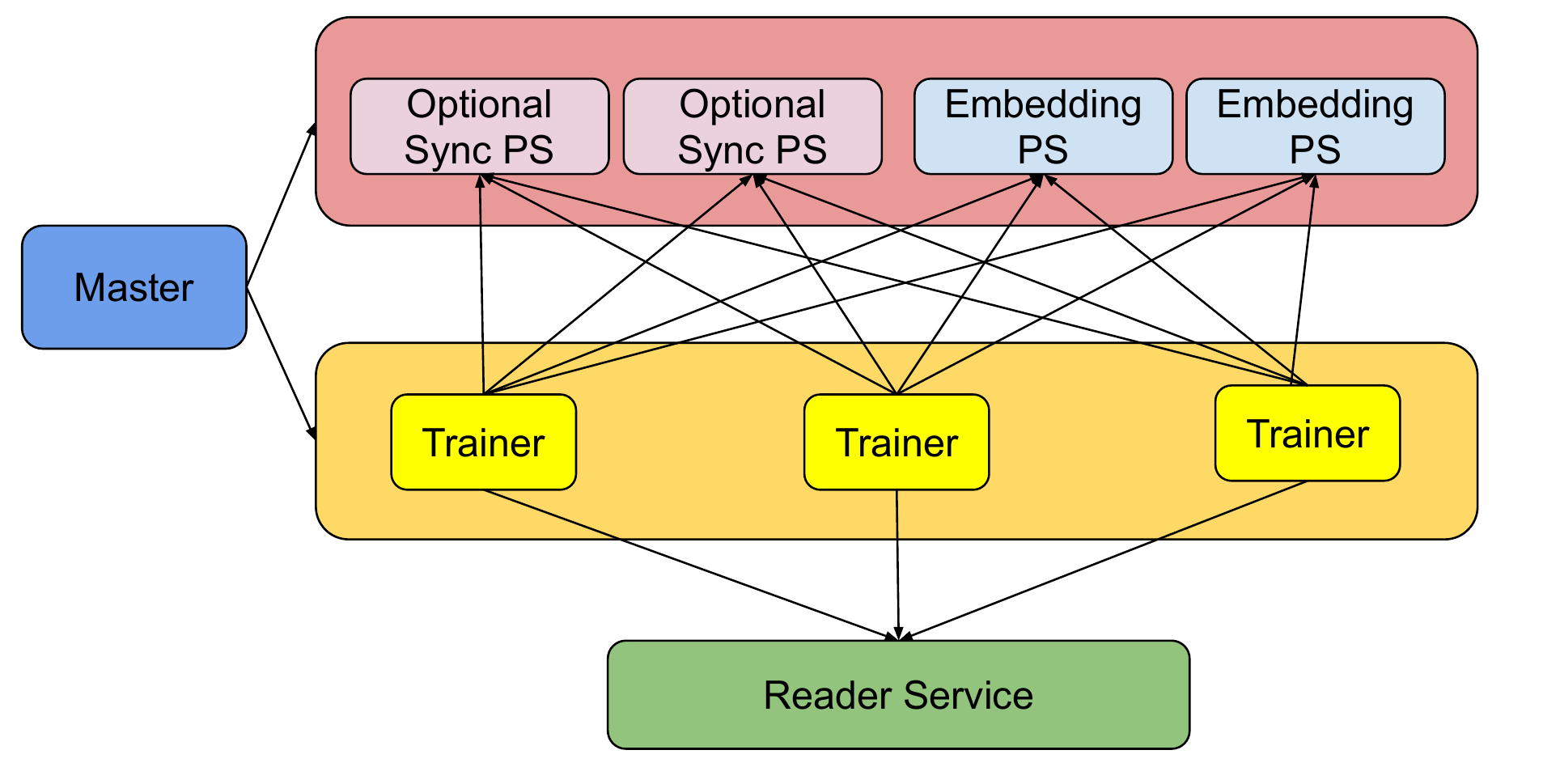}
    \vskip-5pt
    \caption{Overview of our distributed training system. 
    For architectures similar to DLRM with gigantic embedding tables, 
    the embedding tables are served in the embedding PSs and shared by all the trainers,
    and the other parameters are replicated in each trainer.
    }
    \label{fig:system_arch}
\end{figure}
We summary our framework in Fig~\ref{fig:system_arch}. A master machine coordinates the overall training process. There are 3 roles of machines in our framework: trainers, embedding parameter servers (PS), and sync parameter servers (PS).
The master machine assigns different roles to the worker machines and send training plans to them for execution. In the following, we explain the roles of the worker machines and how they jointly train the DLRM.

\subsubsection*{1. Embedding PS and Model Parallelism of the Embedding Table.}
For industry scale problems, a typical DLRM model might contain hundreds of embedding tables with in total billions of rows,  and it easily exceeds the memory of a single machine. To address this issue, we express \emph{model parallelism} on the embedding components:
we partition the embedding tables into smaller shards that fit into the memory of single machines, and serve 
them in machines we call embedding parameter servers(embedding PS). The parameters (embeddings) in the embedding PSs are shared among all the trainers; trainers will communicate the embedding PSs to request and update the embeddings. 
Sometimes, we would like to compute the interactions of several latent embeddings, for example, the attention mechanism of latent embeddings. Those computations are carried out in the embedding PSs to reduce the communication cost. 
As a simple example, the average pooling for several embeddings is computed on the PS directly, and only the final result is returned to the trainer. If one embedding table is partitioned into multiple shards and placed on multiple embedding PSs, we will perform local embedding pooling on each PS for the local shard. The partial pooling results from the shards will be returned to the trainer to perform the overall pooling to get the final results. 

\shadowsync~will automatically ensure the workload of different embedding PSs is distributed evenly. To achieve this,
we perform a quick performance estimation via training a small subset of data before the normal training process. In this phase, 
we profile the cost of embedding lookup and then solve a bin packing problem to distribute the workload (the embedding lookup cost) among the embedding PSs (the available bins) evenly. With this optimization, we are able to ensure that the trainers are not bottlenecked by the shared embedding PSs. 

\subsubsection*{2. Sync PS and Data parallelism of the Other Components.}
On the other hand, the other components (e.g. bottom and top MLPs, feature interaction components) are often small and can fit in the memory of one machine. We then express \emph{data parallelism} for the other components: we replicate those parts of parameters
in each trainer. With that, each trainer will process different batches of examples in parallel, and update its own local copy of the parameters, and then synchronize the obtained (sub-)models from time to time. The synchronization can be conducted in a decentralized manner where trainers aggregate models themselves; there can also be a central machine that receives local copies of models from the trainers and aggregate them. We call this machine a \emph{sync parameter servers}(sync PS).
In such case, since the synchronization is a network heavy operation, letting one PS sync with all the trainers
might make it the bottleneck
Therefore, we allow partitioning the parameters and use multiple sync PSs to sync the parameters. Similar to the embedding PSs, profiling
and optimization are applied to balance the workload of the sync PSs.


\subsubsection*{3. Trainer.}
The trainers execute the training plans. 
Given a batch of data, a trainer computes the bottom feature transformation MLP and
sends the embedding lookup requests to the corresponding embedding PSs.
After all the embedding lookups are returned, the trainer executes the interaction components (except the part computed on the embedding PS) and the prediction MLP to finish the forward pass.
For the backward pass, the trainer computes the gradients for all the parameters. 
The gradients of the embedding parameters are sent back to embedding PS, whereas
the local copy of the MLPs and interaction components are updated directly.

\subsubsection*{4. Reader Service.}
The trainers are connected to a shared reader service,
a distributed system which consumes the raw data in the distributed storage, and then convert the raw data to feature tensors. The reader service system takes the data processing duty, and ensures that the trainers would not be bottlenecked on data reading.

\vskip3pt
Even though we demonstrate our framework with the DLRM, 
it is worth emphasizing \shadowsync~readily generalizes to other deep architectures in all the areas, whenever
the underlying mathematical formulation is an unconstrained optimization problem.

\subsection{Optimization Strategies}\label{sec:optimization}
\begin{figure}[tb]
\begin{subfigure}[t]{0.48\columnwidth}
    \centering
    \includegraphics[width=0.7\columnwidth]{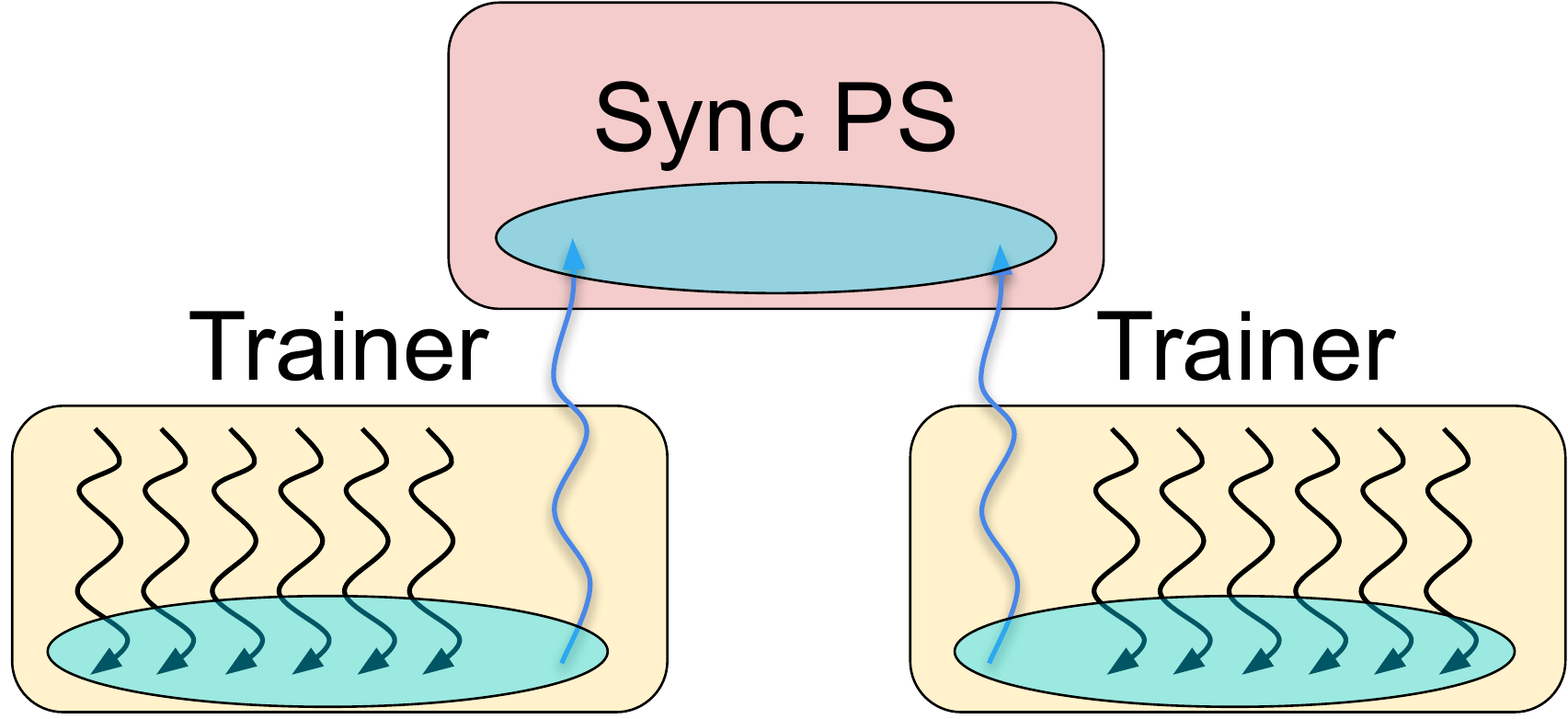}
    \caption{ShadowSync with centralized algorithms. The shadow threads will talk to Sync PSs. There is no interaction among the trainers.}
\end{subfigure}
\hspace{0.01\columnwidth}
\begin{subfigure}[t]{0.49\columnwidth}
    \centering
    \includegraphics[width=0.7\columnwidth]{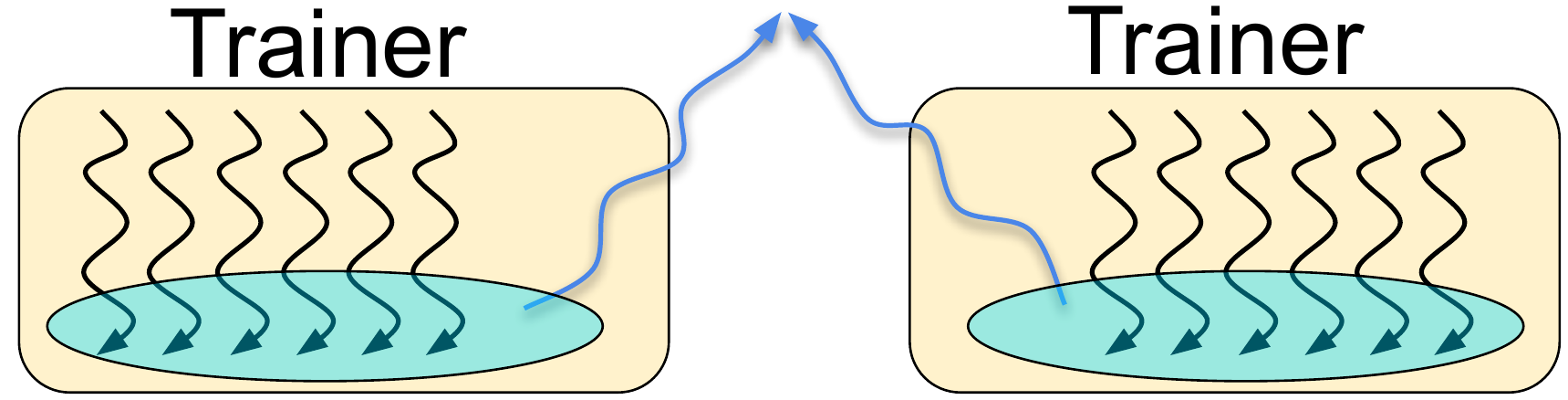}
    \caption{ShadowSync with decentralized algorithms. Sync PSs are absent and the shadow threads will communicate with each other.}
\end{subfigure}
    \vskip-5pt
    \caption{ShadowSync for data parallelism. The black arrows represent worker threads. 
    They update local replica of parameters in the Hogwild manner. The blue arrows represent shadow threads whose job is synchronization.}
    \label{fig:shadowsync}
\end{figure}
To fully utilize the computing resources, all the machines are multi-threaded. On each trainer,
a number of worker threads are processing the example batches in parallel. The simplest parallelization idea is to let each thread process one individual batch of examples. A more complicated mechanism is to explore intra-operation parallelism, so that we can use multiple threads to compute one example batch. This is beyond the scope of this paper, and we will assume that each thread processes one individual batch in this work. The parameter servers also have multiple threads to handle requests sent by trainers in parallel.

Given that we have performed different parallelization strategies for the embedding tables and other components,
we design different optimization strategies for them.


\begin{figure}
  \centering
  \includegraphics[width=0.15\columnwidth]{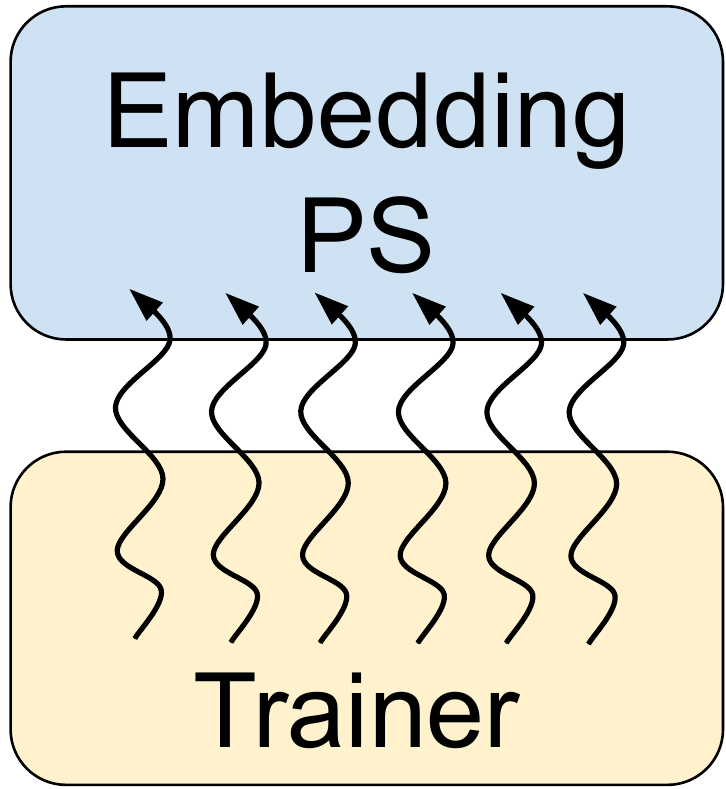}
  \caption{The worker threads optimize the embedding tables using Hogwild.}
  \label{fig:sync_embedding}
\end{figure}
\subsubsection*{Embedding Tables: Hogwild Updates}
The embedding tables are partitioned into many shards and hosted in different embedding PSs. Therefore, there is only one copy of the embedding tables in the whole system.
With that, we optimize the embedding tables using the Hogwild\cite{hogwild} algorithm, see Fig~\ref{fig:sync_embedding}.
In short, this is a lock-free approach. When an embedding PS thread receives one request from a trainer, it reads or writes the embedding table 
without any locking. Different optimizers can be used to update the embeddings, such as Adagrad, Adam, and so on.
All the auxiliary parameters for the optimizers (e.g., the accumulation of the squared gradient for Adagrad) collocate with the actual embeddings on the embedding PSs.
\subsubsection*{Interactions and MLPs: Shadow Synchronization}
For the interaction and MLP components, data parallelism is expressed, so that the parameters for these components are replicated across all the trainers.
Locally, all the worker threads within one trainer access the shared memory space, updating a single copy of the parameters. These accesses are performed in a Hogwild manner as well: the reads and the updates to the local parameters are lock-free.\footnote{
This strategy has broken the Hogwild assumption that the parameter accesses are sparse. 
In our setup,  all the threads are essentially accessing the same parameters simultaneously.
In practice, this strategy works pretty well and provides excellent convergence, see Section~\ref{sec:expr_hogwild} for the experimental results.}
Similarly, they also share the auxiliary parameters of the optimizer.
In order to synchronize the local (sub)-models, in each trainer, 
we create one \emph{shadow} thread independent of the worker threads to carry out the synchronization.
This thread is simply looping the synchronization without any knowledge of the training process.
\shadowsync~is flexible to host almost any kind of synchronization algorithms.
We can use decentralized algorithms like model average\cite{allreduce} and BMUF\cite{bmuf}, and the shadow threads could communicate with each other. Alternatively, if we uses a centralized algorithm like EASGD\cite{easgd}, the shadow threads will sync with the servers; see Fig~\ref{fig:shadowsync} for an illustration.

\vskip6pt
To summarize, we use Hogwild to optimize the model parallelized parameters (the embedding tables), and 
we have two layers of data parallelism in \shadowsync. 
The first layer is the intra-trainer parallelism across the threads, for which we use Hogwild too. On top of it,
the second layer
is the parallelism across the trainers, for which we synchronize using the shadow threads with any algorithm of the users' choice.

This framework has a number of appealing properties.
First,
as we have separated the duty of training and synchronization into different threads,
training is never
stalled by the synchronization need. As a consequence, the huge synchronization overhead is removed. This is confirmed by
our experiments in Section~\ref{sec:expr_easgd}.
Second, our system is capable of expressing different sync algorithms. Both decentralized and centralized algorithms can be hosted by \shadowsync. Similarly, both synchronous and asynchronous versions
of algorithms can be plug into \shadowsync. See Section~\ref{sec:algo}, where we have incorporated the centralized asynchronous EASGD, decentralized synchronous model average into our framework.
Last but important, in the practical realization of our system, the development of synchronization algorithms can be completely separated from training code. This makes the system easy to extend, modify and experiment, without much engineering effort.

\begin{table}[t]
    \centering
    \small
    \begin{tabular}{c|c|c|c|c}
    \toprule
                            & BG/FG     & Hogwild   & Sync/Async    & C/DC  \\ \midrule
    ShadowSync              & \textbf{BG}        & \textbf{Yes}       & \textbf{Flexible}          & \textbf{Flexible}  \\ \hline
    EASGD \cite{easgd}      & FG        & No        & \textbf{Flexible}         & C     \\ \hline
    DC-ASGD \cite{dcasgd}   & FG        & \textbf{Yes}       & Async         & D     \\ \hline
    BMUF \cite{bmuf}        & FG        & No        & Sync          & DC    \\ \hline
    DownpourSGD \cite{distbelief} & FG  & No        & Async         & C     \\ \hline
    ADPSGD \cite{adpsgd}    & FG        & No        & Async         & DC    \\ \hline
    LARS \cite{lars}        & FG        & No        & Sync          & DC    \\ \hline
    SGP \cite{sgp}          & FG        & No        & Async         & DC    \\
    \bottomrule
    \end{tabular}
    \vskip5pt
    \caption{Features of different synchronization approaches. 
    BG/FG means whether sync happens in foreground or background.
    C/DC means whether the algorithm is centralized or decentralized.
    }
    \label{tbl:alg_compare}
\end{table}
Finally, we compares the synchronization behaviors of \shadowsync\xspace with the other existing approaches, 
see Table~\ref{tbl:alg_compare}. 
The BG/FG column states whether the approach performs synchronization in the foreground as part of the training loop, or in the background threads that do not interfere with the training loop. The Hogwild column states whether the approach allows multiple threads in a host to access and update the shared parameters in a lock-free way. The Sync/Async column states whether the approach allows synchronous or asynchronous synchronization protocol. The C/DC captures whether the approach uses centralized (global parameters are hosted in centralized parameter servers) or decentralized strategy. \shadowsync is the only one that performs synchronization in the background, allows lock-free updates in each trainer, and is flexible in terms of the synchronization protocol (sync/async) and the synchronization topology (centralized/decentralized).

%% file: 033-algo.tex
\subsection{ShadowSync Algorithms}
\label{sec:algo}
In this section we present the formal algorithmic description of the ShadowSync concept.
Three representative algorithms under this framework are introduced, which
incorporate the synchronization strategy of EASGD \cite{easgd}, model averaging
(MA)\cite{allreduce}, and BMUF \cite{bmuf} into the execution plan of shadow threads
respectively. We call these algorithms \emph{Shadow EASGD}, \emph{Shadow MA} and
\emph{Shadow BMUF}.

Suppose there are $n$ trainers. Recall there is only one copy of the embedding parameters and $n$ replications of the other parameters in the system. We denote them by $h$ and ${w^{(1)}}_{i=1}^n$, respectively.
Let $D^{(i)}$ denote the data consumed by trainer $i$ and $f$ denote the objective function.
As a unified presentation, \shadowsync~solves the following optimization problem:
\begin{equation}\label{eq:obj}
\min_{w^{(1)}, \ldots, w^{(n)}, \, h} \sum_{i=1}^n f_{D^{(i)}}(w^{(i)}, h) + \lambda R(w^{(1)}, \ldots, w^{(n)}),
\end{equation}
where the regularizer $R$ promotes the consistency across the weight replicas.
It has different forms for different algorithms. For exmample, for 
the model averaging algorithm, $R(w^{(1)}, \ldots, w^{(n)})$ is the indicator function
$\mathds{1}_{  w^{(1)} = \ldots = w^{(n)} }$; 
for EASGD, $R(w^{(1)}, \ldots, w^{(n)}) = \sum_{i=1}^n \norm{w^{(i)} - \widetilde{w}}^2$,
where $\widetilde{w}$ is the auxiliary copy of parameter on the sync PS.

Algorithm~\ref{algo:shadowsync} summarizes the ShadowSync idea.  We first initialize
the embedding tables by $h_0$.  The initialization of MLP and interaction layers $w_0$
are fed to all the trainers. If we use centralized algorithms, the Sync PSs need to be
present and be initialized by $w_0$ too.  The worker threads on each trainer will
optimize their own local weight and the embedding table in the lock-free manner. In
other words, if there are $m$ worker threads spawned per trainer, the embedding $h$
will be updated using $nm$ Hogwild threads across the trainers, and the local copy
$w^{(i)}$ will be updated by $m$ Hogwild threads within trainer $i$.  
For decentralized algorithms, the update of $w^{(i)}$ will involve copies on other trainers, 
whereas for centralized algorithms, $w^{(i)}$ will just sync with $w^{\text{PS}}$. 

Algorithm~\ref{algo:easgd}, \ref{algo:ma}, \ref{algo:bmuf} describe the synchronization
mechanisms of Shadow EASGD, Shadow MA and Shadow BMUF. Contents of worker threads and
initialization that are repeating Algorithm~\ref{algo:shadowsync} are omitted.  For MA,
each trainer will host an extra copy of weights $w^\text{global}$, which is used to
aggregate the training results via \texttt{AllReduce}. Similarly we have auxiliary $w^\text{copy}$
and $w^\text{global}$ for BMUF. To sync, BMUF defines the difference
between the latest averaged model and current $w^\text{global}$ as the descent
direction, then make a step along it. Considering the descent direction as a surrogate
gradient, one can incorporate techniques like momentum update and Nesterov acceleration 
into the updates.
\begin{algorithm}[t]
\SetKwBlock{DoParallel}{trainer $i$ do in parallel with others}{}
\SetKwBlock{DoWorker}{worker threads do in parallel}{}
\SetKwBlock{DoShadow}{shadow thread do}{}
\DontPrintSemicolon
\kwInput{$w_0$, $h_0$}\\
 Init embedding tables on embedding PSs: $h \leftarrow h_0$\;
 (Optional) Init MLP \& interaction params on sync PSs: $w^{\text{PS}} \leftarrow w_0$\;
 \DoParallel{
 Init local MLP and interaction param $w^{(i)} \leftarrow w_0$\;
    \DoWorker{
     \While{data is not all consumed}{
        Update $h$ on embedding PSs\;
        Update local param $w^{(i)}$\;
      }
    }
    \DoShadow{
     \While{data is not all consumed}{
        Sync local param $w^{(i)}$ with Sync PS or other trainers
      }
    }
 }
\caption{ShadowSync Framework}
\label{algo:shadowsync}
\end{algorithm}
\begin{algorithm}[t]
\DontPrintSemicolon
\kwInput{elastic param $\alpha$}\;
\SetKwBlock{DoShadow}{shadow thread do}{}
\DoShadow{
    \While{data is not all consumed}{
        $\wps \leftarrow (1-\alpha) \wps + \alpha w^{(i)}$\;
        $w^{(i)}  \leftarrow (1-\alpha) w^{(i)} + \alpha \wps$\;
    }
}
\caption{Shadow EASGD on Trainer $i$}
\label{algo:easgd}
\end{algorithm}
\begin{algorithm}[t]
\DontPrintSemicolon
\kwInput{elastic param $\alpha$, total number of trainers $n$}\;
\SetKwBlock{DoShadow}{shadow thread do}{}
Init MA global param $w^\text{global} \leftarrow w_0$ \;
\DoShadow{
    \While{data is not all consumed}{
    $w^\text{global} \leftarrow w^{(i)}$ \tcp*{make a copy of local param}
    $w^{\text{global}} \leftarrow \texttt{AllReduce}(w^\text{global}) / n$\;
    $w^{(i)}  \leftarrow (1-\alpha) w^{(i)} + \alpha w^\text{global}$
    }
}
\caption{Shadow MA on Trainer $i$}
\label{algo:ma}
\end{algorithm}
\begin{algorithm}[t]
\DontPrintSemicolon
\kwInput{step size $\eta$, elastic param $\alpha$, total number of trainers $n$}\;
\SetKwBlock{DoShadow}{shadow thread do}{}
Init BMUF global param $w^\text{global}, w^\text{copy} \leftarrow w_0$\;
\DoShadow{
    \While{data is not all consumed}{
        $w^\text{copy} \leftarrow w^{(i)}$ \tcp*{make a copy of local param}
        $w^{\text{copy}} \leftarrow \texttt{AllReduce}(w^\text{copy}) / n$\;
        $w^\text{desc} \leftarrow  w^\text{copy} - w^\text{global}$
        \tcp*{compute descent direction}
        \vskip2pt
        \hskip-4pt\tcc{can do momentum update, Nesterov acceleration etc.}
        $w^\text{global} \leftarrow w^\text{global} + \eta w^\text{desc}$\;
        \vskip2pt
        $w^{(i)}  \leftarrow (1-\alpha) w^{(i)} + \alpha w^\text{global}$
    }
}
\caption{Shadow BMUF on Trainer $i$}
\label{algo:bmuf}
\end{algorithm}
The sync update of Shadow EASGD is essentially the same as original EASGD. Given elastic
parameter $\alpha$, it will do convex interpolation between $w^\text{PS}$ and $w^{(i)}$.
Note that the interpolation is asymmetric: $w^{(i)}$ and $w^\text{PS}$ are not equal
after this update. Intuitively, the sync PS is also talking with other trainers, and the trainer did not stop local training during its synchronization with the PS, so that both of them would like to trust their own copy of
weights. Similar interpolation is happening for both Shadow MA and Shadow BMUF. This is
a major modification from the original methods. We observe that
this modification is important for the model quality in the ShadowSync setting. 
Take MA for example, the
\texttt{AllReduce} primitive is time-consuming and the worker threads would have consumed a
fair amount of data in the \texttt{AllReduce} period. If we directly copy the averaged
weight $w^\texttt{global}$ back, we will lose the updates to the local parameter replicas 
when the background synchronization is happening in parallel.

%% file: 04-experiments.tex
\section{Experiments}\label{sec:expr}
We conducted numerical experiments on training our production DLRM model for
click-through-rate (CTR) prediction tasks to demonstrate the effectivness of \shadowsync. 
All the experiments were using anonymized real data.
Our production model has total size approximately 200GB, 
where the interaction and MLP components are roughly 50mb. Due to privacy issues, 
the other detailed description of specific datasets, tasks and model architectures will
be omitted in this paper, yet we will report the sizes of datasets when presenting the
experiments. 

We implement the EASGD, BMUF and MA under \shadowsync, which we refer to
as \easgd, \bmuf,  and \allreduce. To compare with the classic 
foreground synchronization approaches, we implement the foreground asynchronous EASGD, 
where the worker threads synchronize with the PS every $k$ iterations. We refer to this approach as \freasgd,
and the value $k$ as the sync gap.  Similarly, we implement \frma and \frbmuf.

\paragraph{Section Organization.} 
Section~\ref{sec:expr_easgd} compares our background synchronization approach
with the foreground approach by comparing \easgd with \freasgd, and similarly
for MA and BMUF approaches respectively. We have validated that \shadowsync is favorable
by examining the metrics we introduces below.

Next, Section~\ref{sec:expr_shadow} focuses on the comparison of \easgd, \bmuf and \allreduce
within the ShadowSync framework. \bmuf and \allreduce are typical \emph{decentralized}
algorithms -- the usage of sync PSs is eliminated. Those lightweight
algorithms are suitable for scenarios where the computation resource is on a tight
budget. We are thus curious about whether the performance of \bmuf and
\allreduce are on par with \easgd.
Finally, recall that \shadowsync has an extra layer of intra-trainer parallelism: the
worker threads are runing Hogwild updates to the local parameters (c.f. Section~\ref{sec:optimization}).
Section~\ref{sec:expr_hogwild} provide a justification for such parallelism and the choice of
24 Hogwild worker threads in the setup.

\paragraph{Evaluation Metrics.} We compare the algorithms in three aspects: \emph{training throughput},
\emph{model quality}, and \emph{resource efficiency}. Note that as explained in Section~\ref{sec:intro}, for all the experiments
we use one-pass training to prevent overfitting. 
We emphasize that under this setting, the training throughput directly reflects the training speed.  
\begin{itemize}
    \item  The training throughtput is measured by the examples-per-second metric defined below.
\begin{definition}[Examples Per Second]
  \label{def:eps}
We define Examples Per Second (\texttt{EPS}) as the average number of examples per second processed by the distributed training system.
\end{definition}
    \item We measure the model quality by the \emph{normalized entropy} (NE) metric introduced in \cite{practical}. 
    In short,
    it is the cross entropy loss value of our predictor $\yhat$ normalized by the entropy of the empirical CTR distribution:
    \begin{equation}
        \text{NE}(\yhat) = \frac{- \frac{1}{N} \sum_{i=1}^N  ( y_i \log \yhat_i + (1-y_i) \log (1 - \yhat_i) ) }{-\{p\log p + (1-p) \log(1-p)\}},
    \end{equation}
    where $p$ is the empirical average CTR, $N$ is the total number of examples, $\yhat_i$ is the predicted probability of click, and $y_i \in \{0, 1\}$ is the ground truth. The smaller the NE value is, the better is the prediction made
    by the model.
    \cite{practical} has compared NE with other common metrics for CTR prediction e.g. AUC, and concluded NE is preferred. We refer the readers to \cite{practical} for details.
  
    \vskip3pt 
    In our experiments, we investigate the NE of the final output model on test dataset, named \textbf{Eval NE}.
    In addition, we also investigate a variant that is calculated during the training process:
    in each iteration $t$, after we update the model parameters, we predict the CTR of data
    in the next iteration $t+1$ and calculate the NE value. Finally, we report the final average value of them
    as the \textbf{Train NE}. We remark that both Train NE and Eval NE are measuring the \emph{prediction} performance of our models, and both are used as important evaluation metrics for our production models. 
    
    \item In practice, it can happen that several approaches have similar quality and throughput performance. In those cases, we
    shall consider the approach that has highest computing resource efficiency, i.e use as few machines as we can. 
\end{itemize}

\paragraph{Experimental Setup.}After the training ends,
the embedding $h$ and the weight replica on the first trainer $w^{(1)}$ are returned as the
output model (this is for simplicity, an alternative is to return the average of all the weight replicas).
For the sake of fair comparison, all the hyper-parameters like elastic parameter and
learning rate were set as the same as in the production setting for all the algorithms.
The hardware configurations are identical and consistent across all the experiments. All
the trainers and PSs use Intel 20-core 2GHz processor, with hyperthreading enabled
(40 hyperthreads in total).  We set 24 worker threads per trainer and 38 worker threads per parameter
server. For network, we use 25Gbit Ethernet.


\begin{table}[t]
    \centering
     \small
    \begin{tabular}{ c|c|c|c}
    \toprule
           & Sync Gap & Train NE & Eval NE  \\ \midrule
    \easgd & 5.21 & \textbf{0.78926} & \textbf{0.78451}  \\ \hline
    \multirow{4}{*}{\freasgd}
     & 5   & 0.78942 & 0.78483  \\ \cline{2-4}
     & 10  & 0.78937 & 0.78508  \\ \cline{2-4}
     & 30  & 0.78942 & 0.78523  \\ \cline{2-4}
     & 100 & 0.78969 & 0.78531 \\
    \bottomrule
    \end{tabular}
    \vskip5pt
    \caption{\small Model quality of the EASGD methods for training \fbdata{1} using 11 trainers. \easgd outperforms
    \freasgd.}
    \label{tbl:easgd_tr}
\end{table}

\subsection{Shadow vs Foreground}\label{sec:expr_easgd}
\subsubsection{\easgd vs \freasgd on A Single Instance}
We first investigate the performance of \easgd and \freasgd
on a single problem instance.
We are interested in (i) finding the best sync gap $k$ for \freasgd, 
(ii) understanding how the sync gap influences NE, and 
(iii) investigating the average sync gap
of \easgd as we cannot explicitly control it.

We apply them to training our DLRM on \fbdata{1}. 
This dataset contains $48,727,971,625$ training examples and \newline
$1,001,887,500$ testing examples. For \freasgd, we tested 4 values for 
the sync gap $k$ : $5, 10, 30$, and $100$.  
For \easgd, we calculate the \emph{average sync gap}
using a few metrics measured during training as the following:
\[
\begin{aligned}
    &\text{avg sync gap}  = \frac{\text{num of iterations trained per sec}}{\text{num of EASGD syncs per sec}} \\
    &\hskip3pt = \frac{\eps/\text{batch size}}{ \text{sync PSs network usage per sec}/\text{size of param $w$}}.
\end{aligned}
\]
The experiment was carried out with 11 trainers, 12 embedding PSs and 1
sync PS. Table~\ref{tbl:easgd_tr} reports the results.
Table~\ref{tbl:easgd_tr} shows that the Eval NE of \freasgd increases
as the sync gap goes up, and the lowest Eval NE is achieve when the sync gap is 5.
The Train NE of \freasgd does not show any pattern correlated with the sync gap. 
\easgd outperforms \freasgd for both Train NE and Eval NE.
The average sync gap of \easgd is 5.21, very close to 5. 

In the next subsection, we will compare the scalability of \easgd and \freasgd. 
We will use sync gap $5$ and $30$ for \freasgd, and denote them by \freasgdfive and \freasgdlarge,
respectively.

\subsubsection{\easgd vs \freasgd: Scalability Study}
\begin{figure*}[t]
    \centering
    \includegraphics[width=0.4\textwidth]{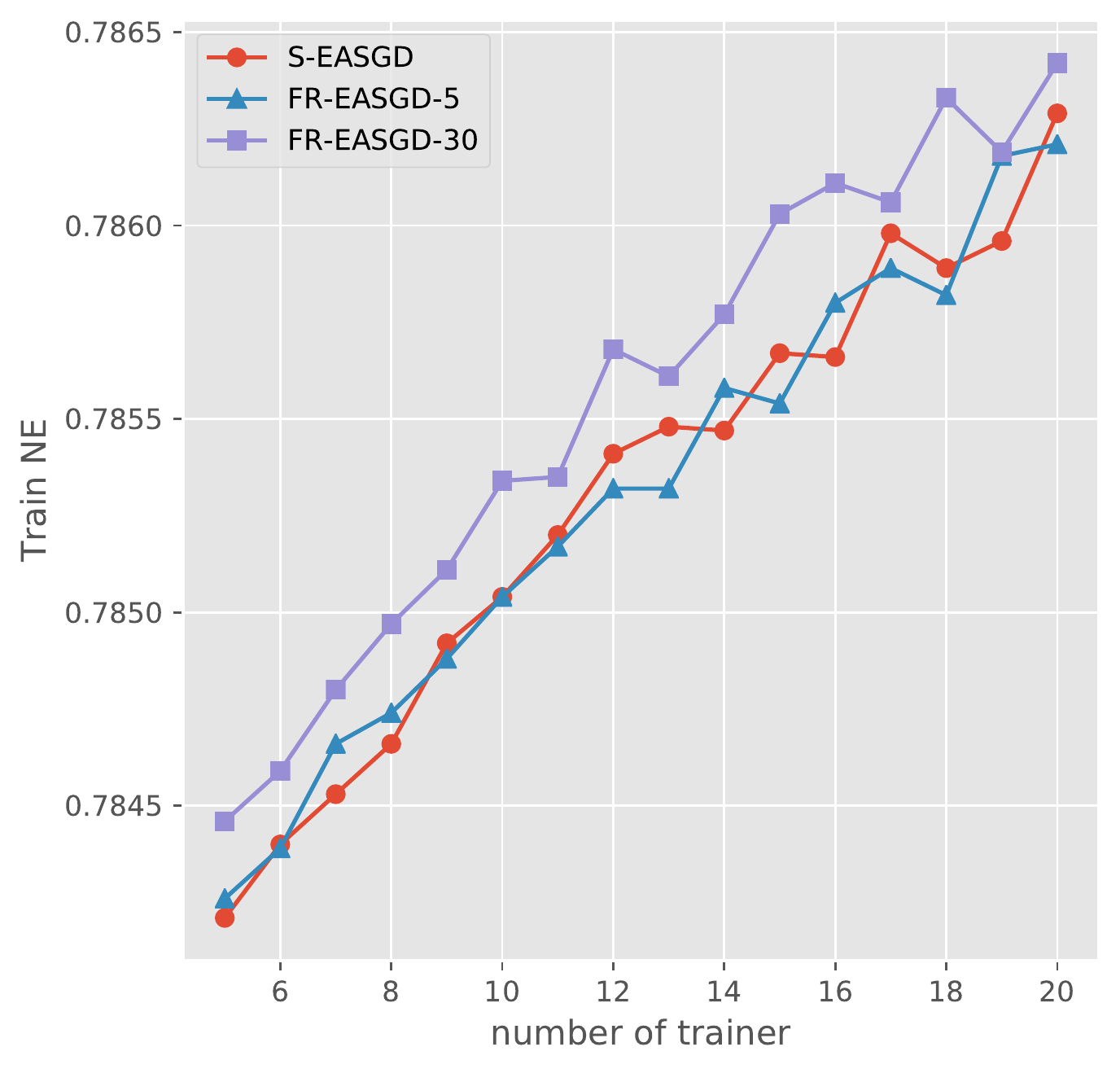}
    \includegraphics[width=0.4\textwidth]{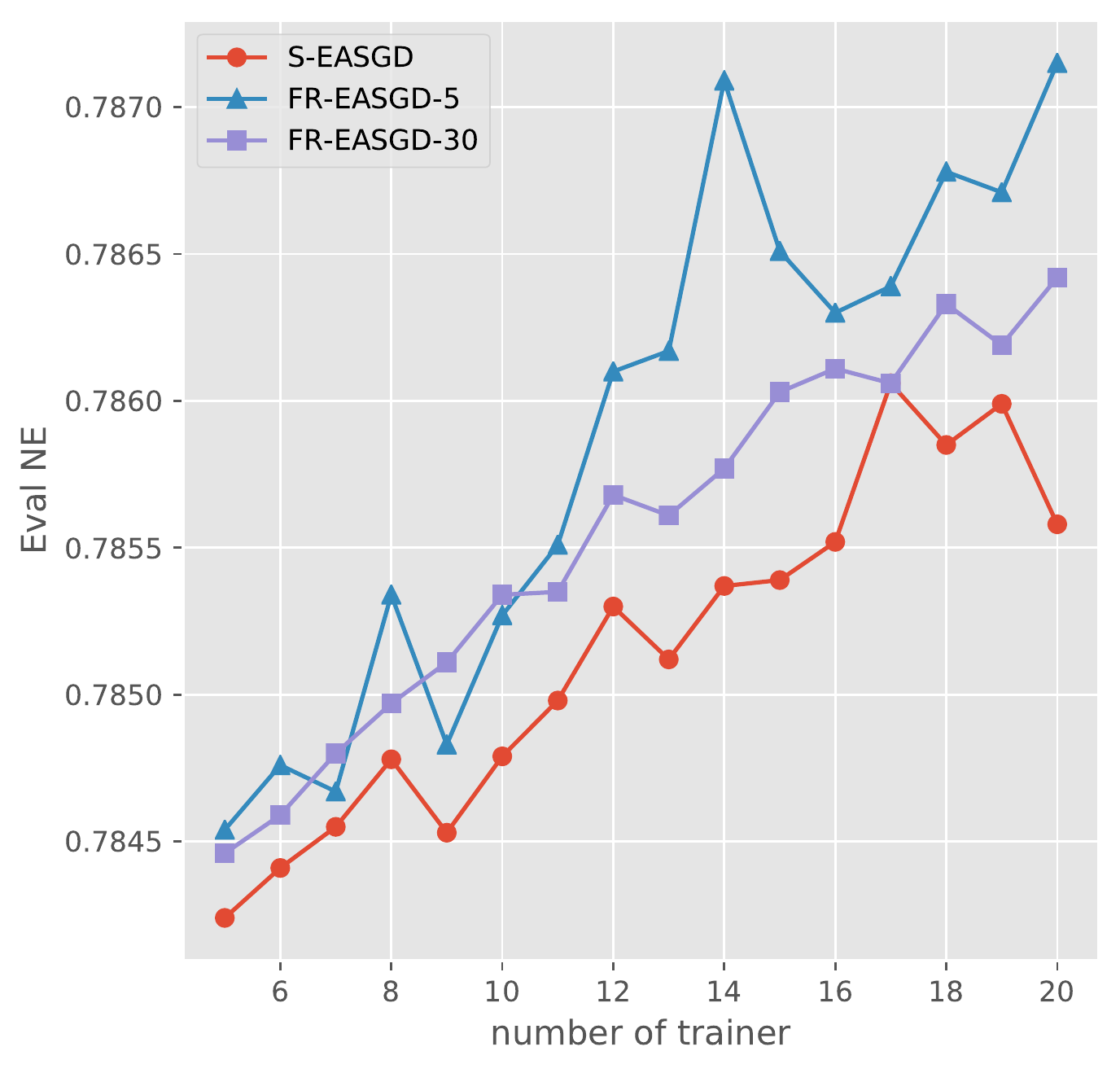}
    \includegraphics[width=0.4\textwidth]{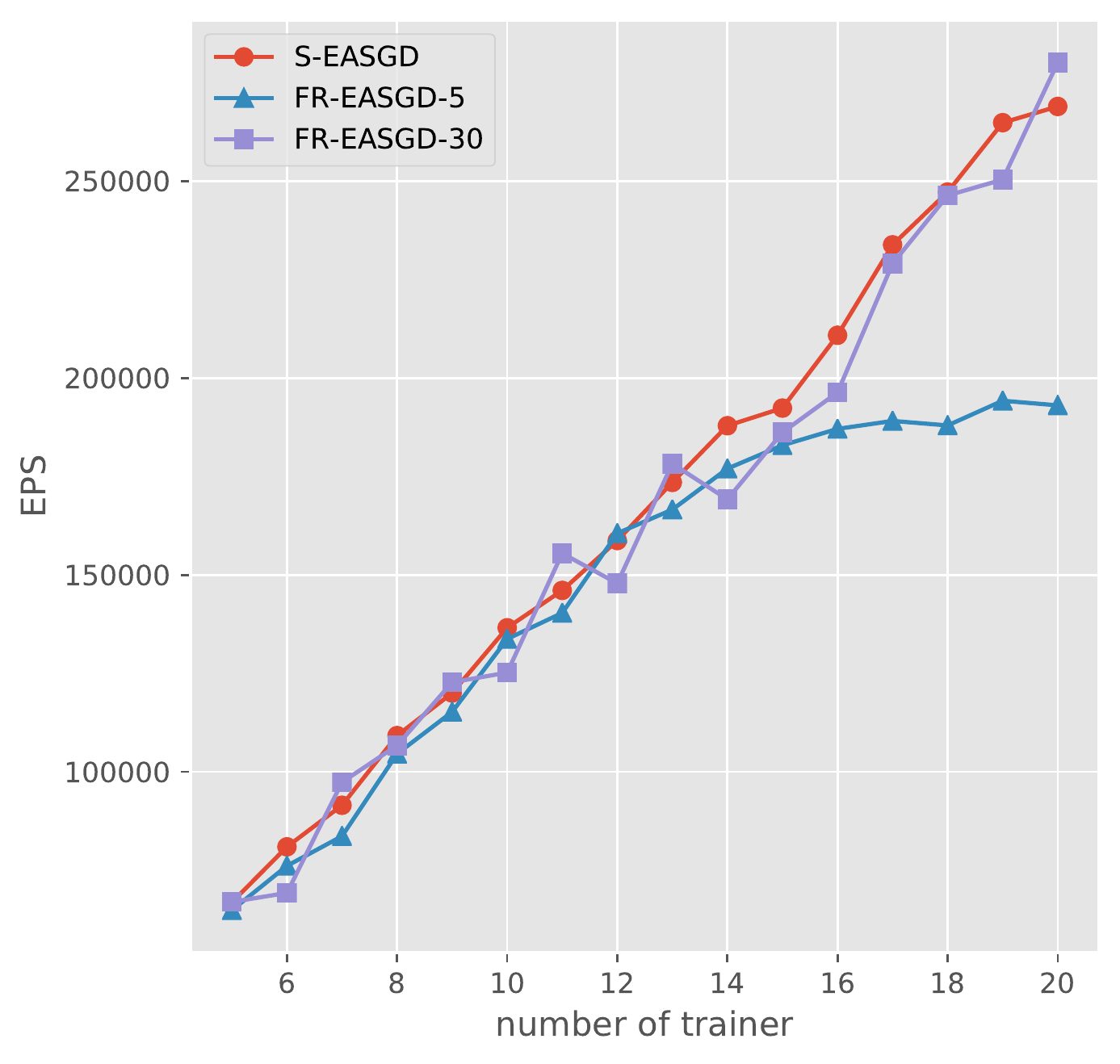}
    \includegraphics[width=0.4\textwidth]{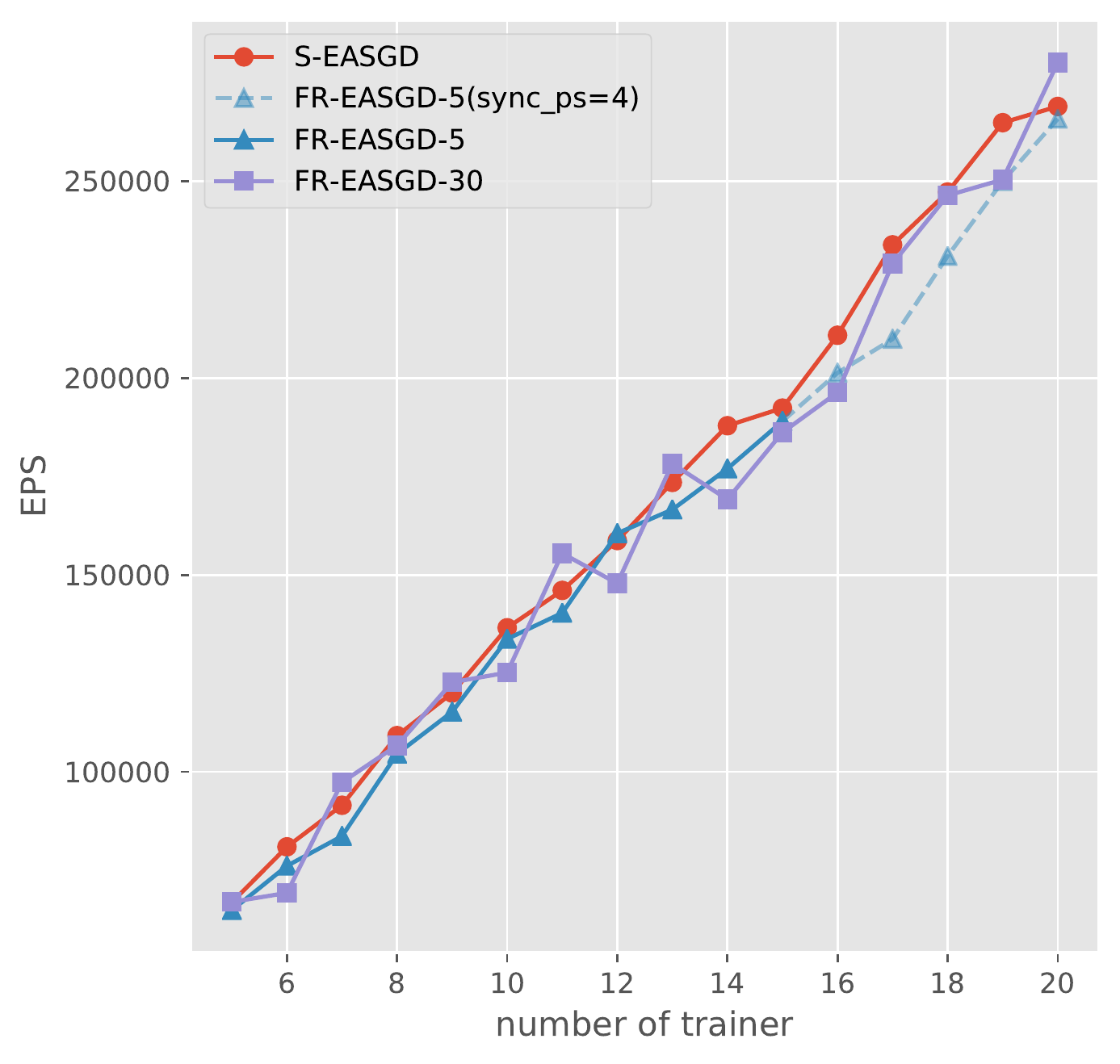}
    \vskip-5pt\caption{The scaling behavior of \easgd and \freasgd for
    training \fbdata{2}.
    }
    \label{fig:new_scala_easgd}
\end{figure*}
\begin{figure*}[t]
    \centering
    \includegraphics[width=0.324\textwidth]{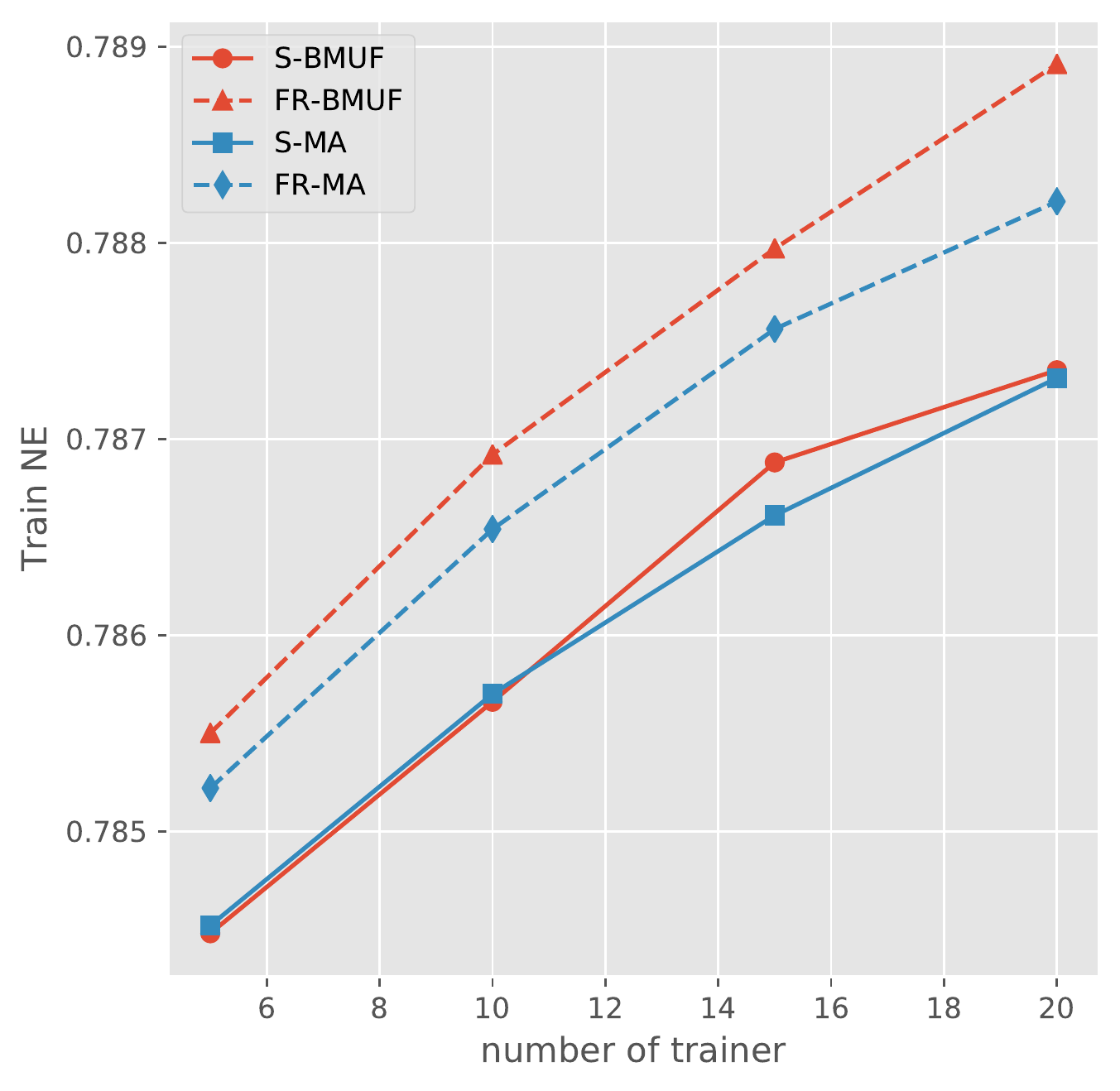}
    \includegraphics[width=0.324\textwidth]{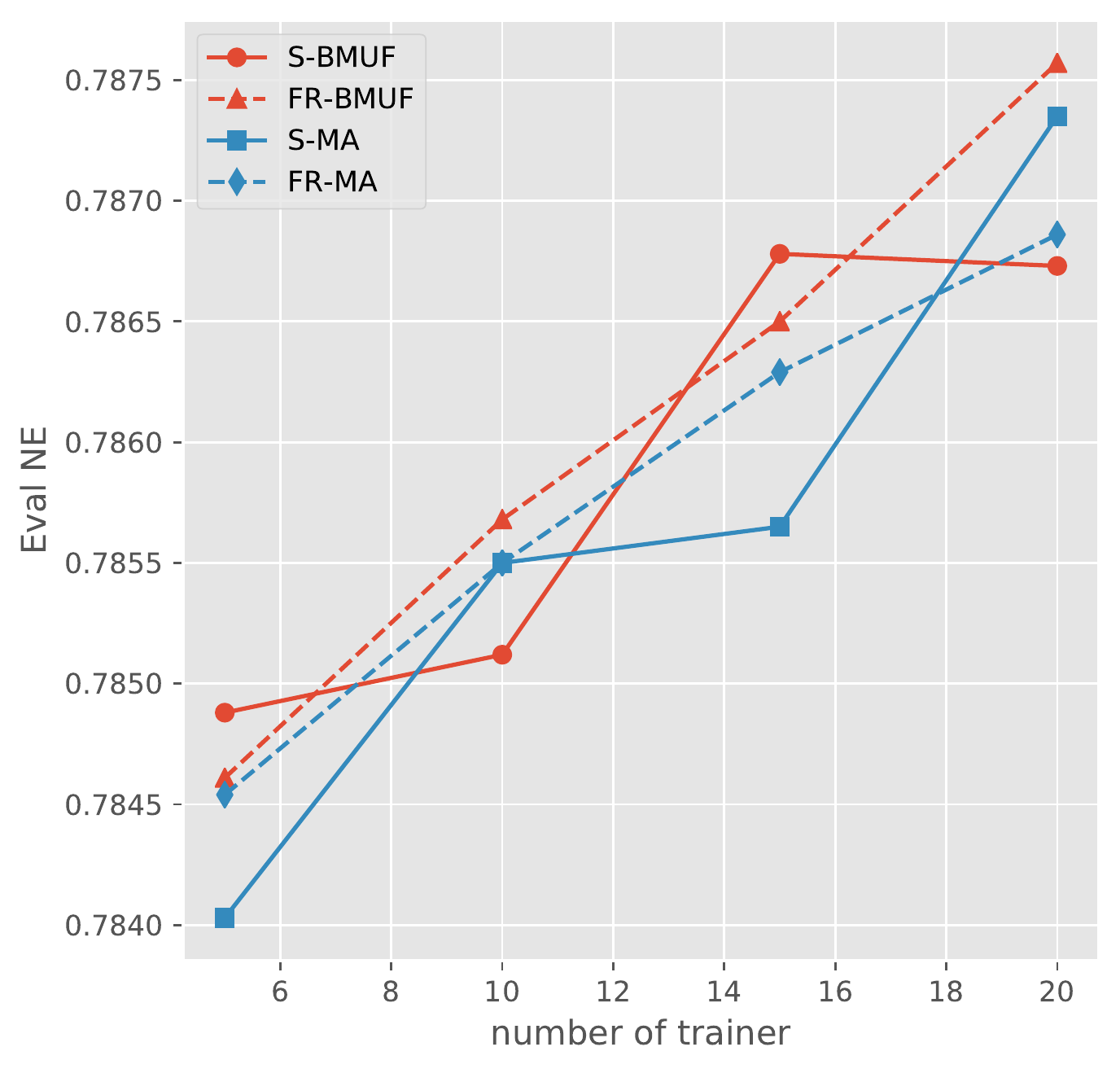}
    \includegraphics[width=0.324\textwidth]{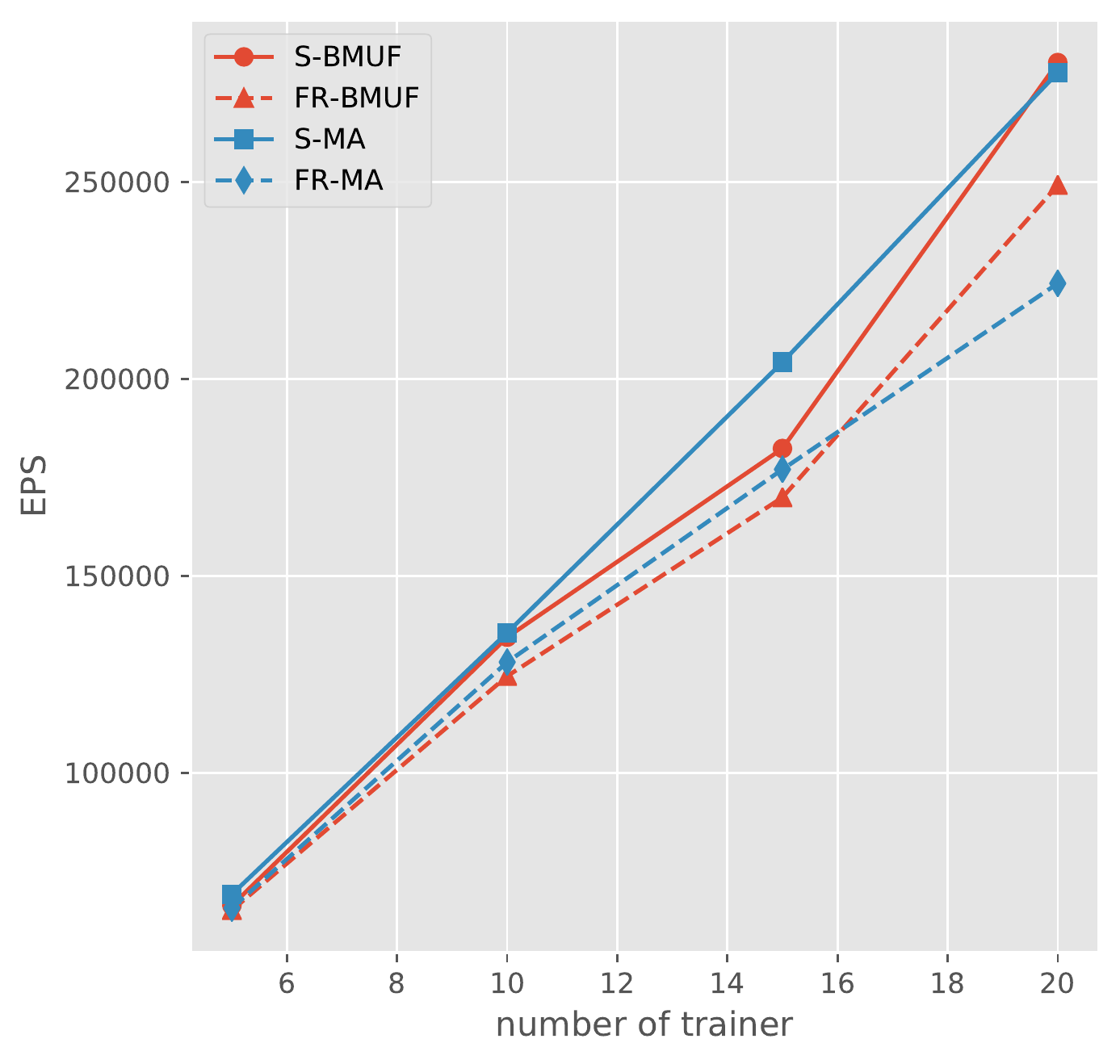}
    \caption{The scaling behavior of MA and BMUF approaches for training \fbdata{2}. The ShadowSync versions achieve lower NE.}
    \label{fig:fr_bmuf_ma}
\end{figure*}
\begin{figure*}[t]
  \centering
    \includegraphics[width=0.324\textwidth]{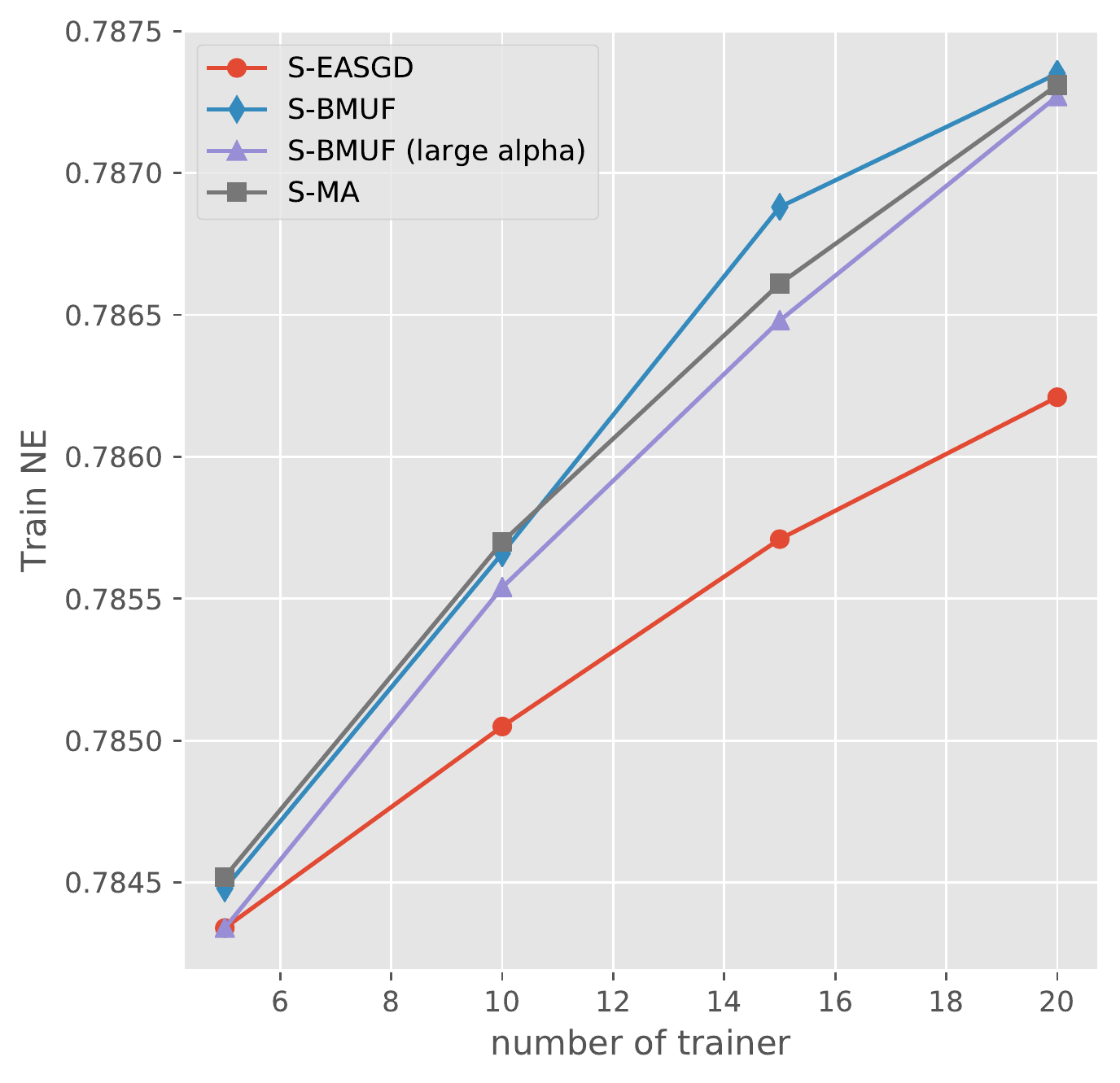}
    \includegraphics[width=0.324\textwidth]{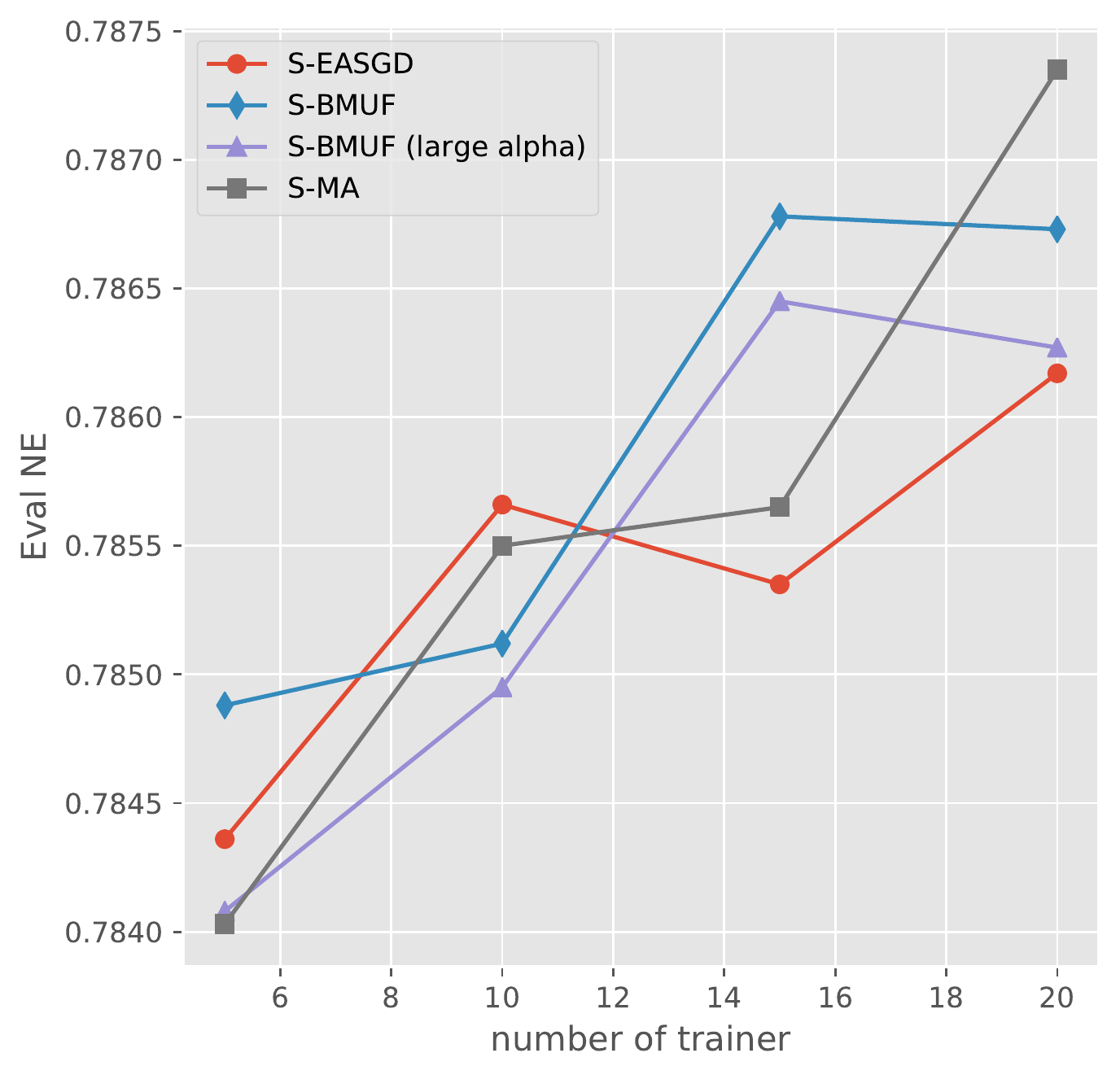}
    \includegraphics[width=0.324\textwidth]{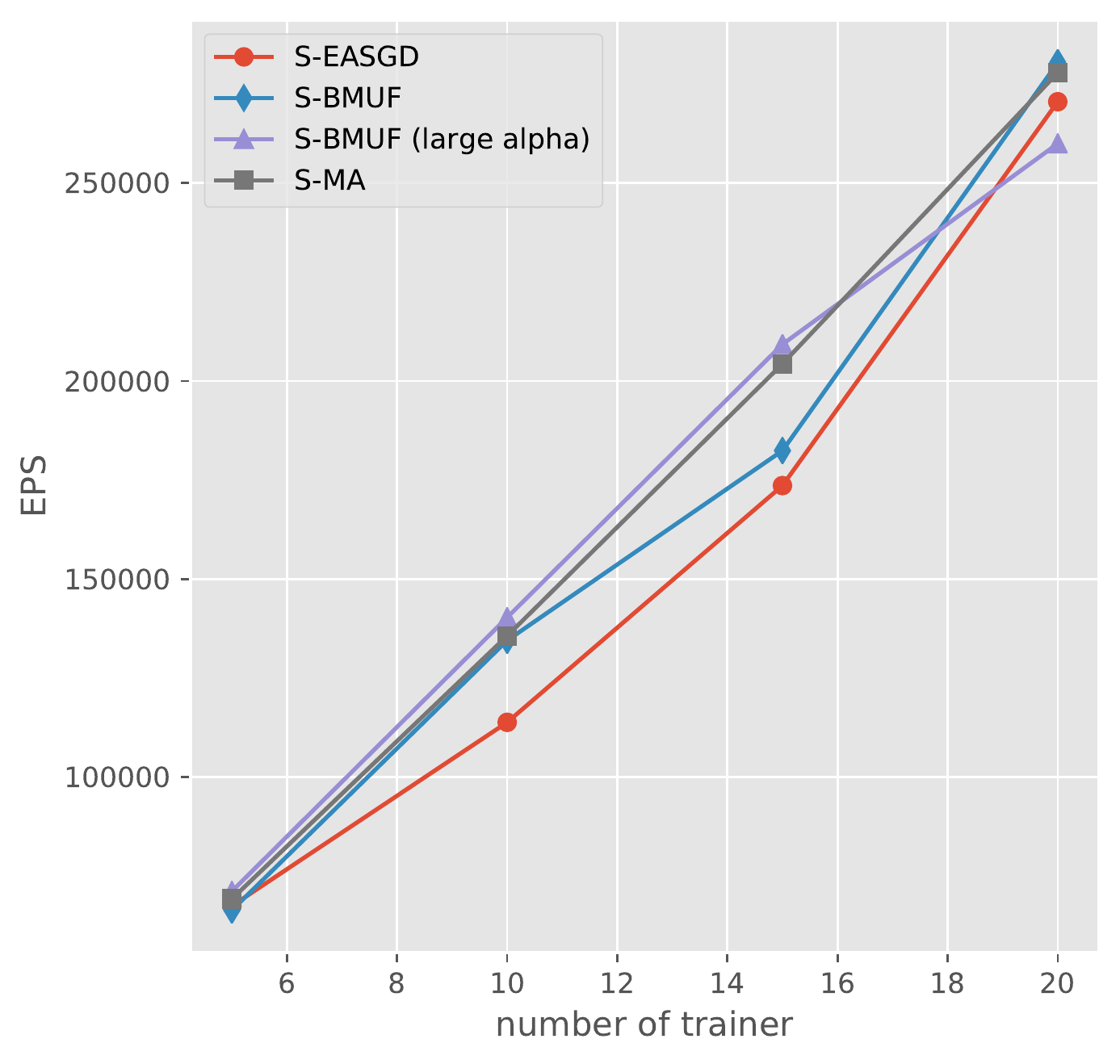}
    \vskip-5pt\caption{Performance of \easgd (centralized), \bmuf (decentralized) and \allreduce (decentralized) for training \fbdata{2}.}
    \label{fig:shadow}
\end{figure*}
\emph{Scalability} is the crucial property of distributed training and the central topic of
our discussion. As we scale up our training by adding more trainers, 
a scalable framework (and associated algorithm) will have
the \eps grow linearly as the number of trainers,
and keep the NE increase small and tolerable \footnote{It is natural to
observe NE increase when one scales up the training.}.

To explore the scaling behavior, we apply \easgd, \freasgdfive and \freasgdlarge to
training DLRM on \fbdata{2}, which contains 3,762,344,639 training examples and 2,369,568,296 testing
examples. We vary the number of trainers from 5 to 20.
 We fix the number of sync PSs to be 2, and
overspecify the number of embedding PSs to be the
same as trainers. This is for demonstration purpose so that the training would not be bottlenecked by the embedding PS.  

\begin{table}[t]
    \centering
    \small
    \begin{tabular}{ c | c|c|c|c}
    \toprule
    \multicolumn{2}{c|}{} & \easgd & \freasgd{}-5 & \freasgd{}-30 \\ \midrule
    \multirow{2}{*}{10 T}
    & Train & \textbf{0.084\%}& 0.099\% & 0.096\% \\ \cline{2-5}
    & Eval & \textbf{0.062\%} & 0.093\%& 0.112\%  \\ \hline
    \multirow{2}{*}{20 T}
    & Train & 0.230\% & 0.249\% & \textbf{0.210\%} \\ \cline{2-5}
    & Eval   & \textbf{0.177\%}& 0.333\% & 0.250\%  \\
    \bottomrule
    \end{tabular}
    \vskip5pt
    \caption{\small The relative NE increase of EASGD approaches when the number of trainers are 10 and 20,
    comparing with the 5-trainer result. The results are reported as for training \fbdata{2}.}
    \label{tbl:easgd_NE_perc}
\end{table}
Fig~\ref{fig:new_scala_easgd} reports the results. 
The 1st and 2nd panels plot compare the Train and Eval NE for all the methods.
The Train NE gently increases in comparable speed for all three methods,
where \easgd and \freasgdlarge are comparable and outperform \freasgdlarge.
Regarding the Eval NE, \easgd achieve lowest NE value for all the number of trainers.
\freasgdfive is not stable in this case.
Setting the 5-trainer case as the base,
we also report the relative increase\footnote{The relative increase of NE is defined as
$(\text{NE}_\text{new} - \text{NE}_\text{old})/ \text{NE}_\text{old}$.}
of NEs when the number of trainers increase to $10$ and $20$.
The results are summarized in Table~\ref{tbl:easgd_NE_perc}.
Clearly, \easgd enjoys the overall mildest NE increase.

The 3rd panel of Fig~\ref{fig:new_scala_easgd} plots \eps as a function of number of trainers $n$.
Both \easgd and \freasgdlarge achieve linear \eps growth. Yet for
\freasgdfive, the \eps almost stopped increasing after the $n$ increased to 14. 
We investigated the hardware utilization
and identified the sync PSs as the bottleneck.
When more trainers are added into training, the network bandwidths of the
sync PSs will saturate at certain point.
Since the synchronization is foreground and integrated
into the training loop for \freasgd, 
the network bandwidth needs of it grow as 24x (we have 24 hogwild threads per each trainer)
compared with\easgd.  Therefore, the sync PSs can easily get saturated especially when
the sync gap is small (so that concurrent synchronization are more likely to happen).
To obtain the linear growth of \eps, we have to increase the number of sync PSs to 4 for \freasgdfive, see the 4th panel.
We also calculated the average sync gap of \easgd as before. For runs with 15 - 20 trainers, the numbers are 8.60, 8.76, 10.43, 10.93, 11.95, and 12.48, which means 
means we synchronize more frequently than \freasgdlarge.

This experiment reveals one \textbf{powerful} strength of \shadowsync framework:
it is able to achieve high throughput and high sync frequency simultaneously.
Instead, the foreground approach needs to sacrifice either the throughput (e.g. \freasgdfive)
or the sync frequency (e.g. \freasgdlarge) which might hurt the model quality, as illustrated in the previous experiment.

\subsubsection{\allreduce vs \frma and \bmuf vs \frbmuf}
\label{sec:expr_bmuf_ma}
We present a similar but simplified scalability experiment for BMUF
and MA type of algorithms on the same dataset. The number are trainers
are set to 5, 10, 15 and 20.
The sync rate of \frbmuf and \frma are set to be 1 per minute
\footnote{The AllReduce primitive is time-consuming so that we decided to
set the sync rate based on time rather than the number of iterations.}.
As reported in Fig~\ref{fig:fr_bmuf_ma},
the performance of ShadowSync algorithms
are comparable and superior to the foreground variants.
Regarding the \eps, since there is no PS involved in the synchronization,
all the algorithms can scale linearly.
The average sync rate of \bmuf is 2 per minute for
5 trainers and 0.8 per minute for 20 trainers. For \allreduce, the numbers were
2.9 and 1.0. 

\subsection{ShadowSync: Centralized vs Decentralized}
\label{sec:expr_shadow}
One shortcoming of the centralized methods is that it requires extra machines as the center hubs
for synchronization purpose only\footnote{one can collocate the trainer and sync PS but this machine
will become the bottleneck and slow down the training.}.
In contrast, for \emph{decentralized} algorithms, the synchronization happens across trainers directly.
There has been an increasing body of research work that focus on developing
decentralized algorithms for saving computing resources.
Therefore, we are interested in comparing \easgd, 
a typical centralized algorithm, with \bmuf and \allreduce,
the two decentralized methods under \shadowsync that suits
users with limited computation budget.

We then check their performance for training \fbdata{2} with 5, 10, 15 and 20 trainers.
We observe that \bmuf tends to update the model more conservatively than \allreduce: it would
make a step towards to average model rather than taking it directly. In light of this, we hypothesized \bmuf will converge slower than \allreduce. Hence, in addition to the standard $\alpha$ used before, we tested a larger value for it to make more aggressive parameter sharing, and report both results.  The other settings are the same as in the previous experiments. See Figure~\ref{fig:shadow} for the results.
Increasing $\alpha$ does improve the performance of \bmuf. \easgd has best Traine NE, followed by \bmuf with larger elastic parameter. The performance on Eval NE is mixed, none of those algorithm stands out clearly.
To summarize, our experiments suggest that \bmuf and \allreduce has the potential to perform comparably good as \easgd.

\subsection{Intra-Trainer Hogwild Parallelism}
\label{sec:expr_hogwild}
\begin{figure}[t]
    \centering
    \includegraphics[width=0.4\columnwidth]{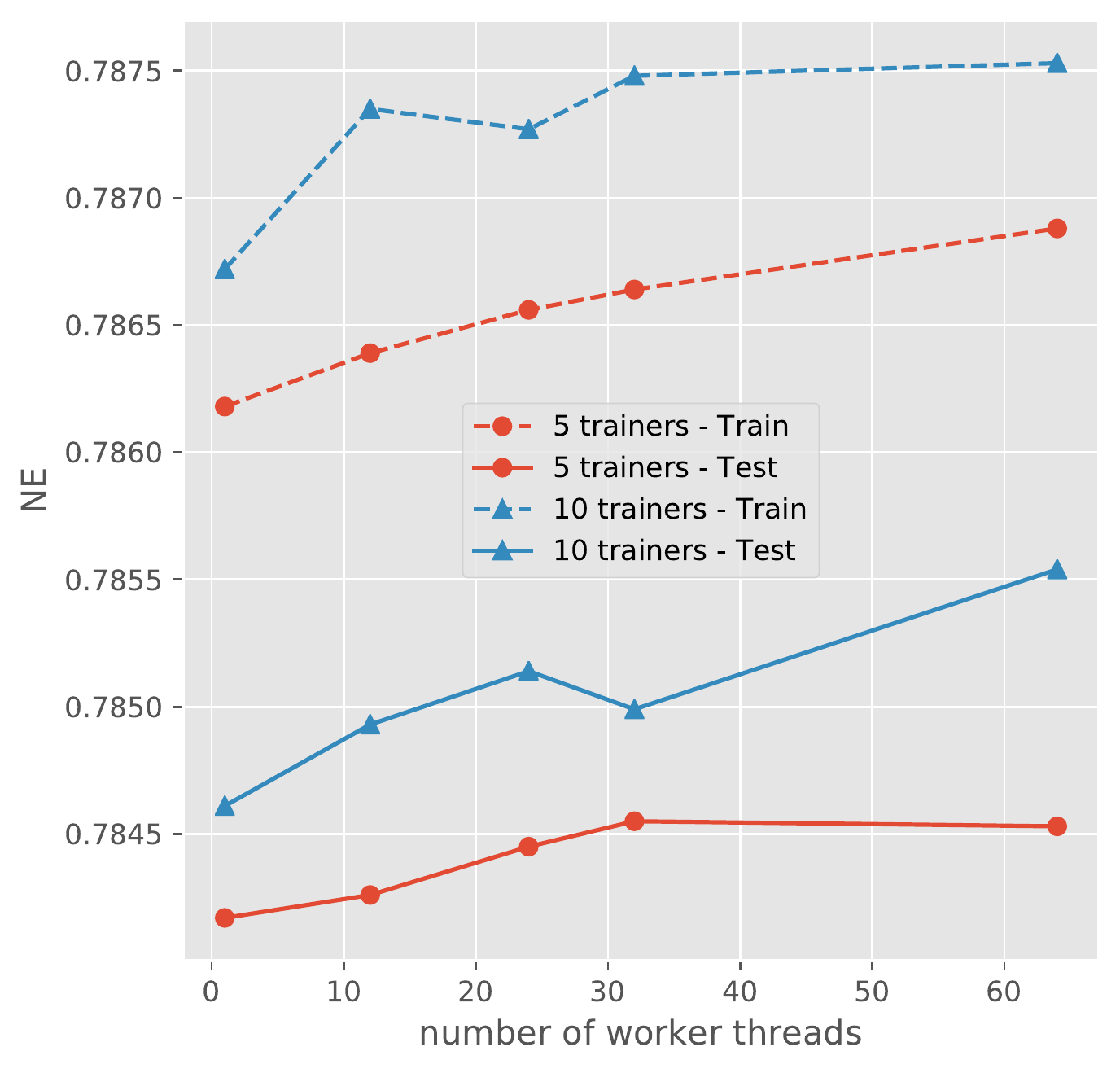}
    \includegraphics[width=0.4\columnwidth]{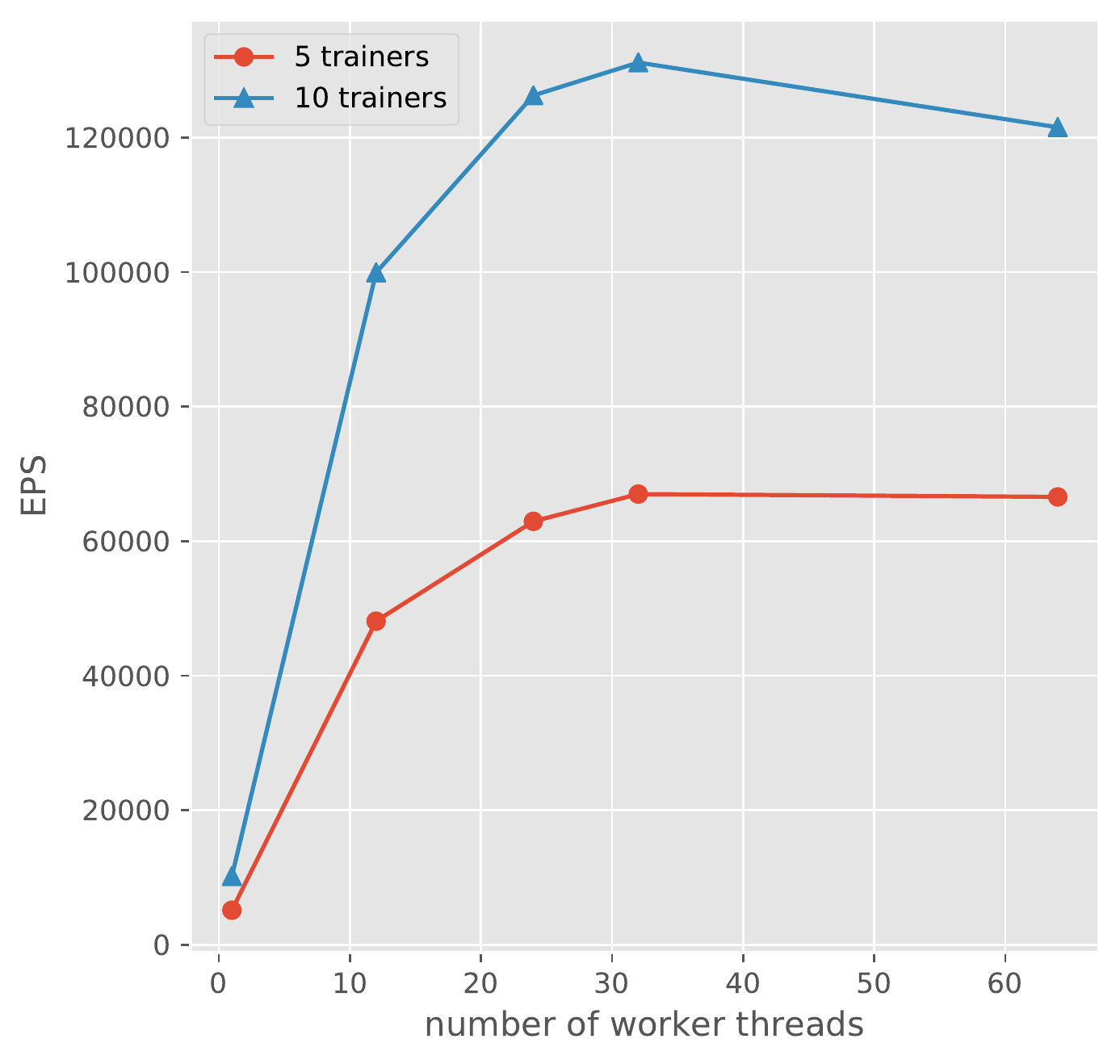}
    \caption{Performance of \easgd with varying number of worker threads for training \fbdata{3}.}
    \label{fig:hogwild}
\end{figure}
Finally, we justify the usage intra-trainer Hogwild prallelism in our framework, and
the choice of 24 worker threads throughout our experiments. 
We apply \easgd to training \fbdata{3} which contains 1,967,190,757 training samples and 4,709,234,620 test samples.
We test under 2 settings: the first one uses 5 trainers, 1 sync PS and 4 embedding PSs; the second one uses 10 trainers, 1 sync PS and 6 embedding PSs.
Under each setting, we run \easgd with 1, 12, 24, 32, and 64 worker threads, respectively. 

Results are shown in Figure~\ref{fig:hogwild}.
The left panel plots Train Ne and Eval NE versus the number of worker threads, and the right panels plots \eps. As expected, the NE increases as the number of worker threads grows, yet such increase is relatively mild compared to the huge \eps gain shown in the right panel. This validates the usage of Hogwild parallelism
in \shadowsync.
The right panel also shows the \eps almost stops growing when the number of worker threads reaches 24, in both settings. We find that the trainers became the bottleneck in those cases, as the memory bandwidth is saturated (the interaction components are memory bandwidth demanding). This justifies our choice of $24$ worker threads.

%% file: 05-conclusion.tex
\section{Conclusion}\label{sec:conc}
We present a distributed training framework, \shadowsync, that allows us to 
train large models with big data. We express model parallelism on the large embedding tables
that do not fit in the memory of a single host; and express intra-trainer and inter-trainer data parallelism
to increase the overall throughput of the system. We have proposed the new idea of isolating synchronization from training and performing it in the background. It allows us to scale \eps linearly
without being bottlenecked by synchronization. The experimental results show that the \shadowsync algorithms have better model quality compared to their foreground counterparts. Moreover, \shadowsync can accomplish linear \eps scaling with fewer machines. This shows that \shadowsync is indeed favorable compared with the foreground algorithms.


%